\documentclass[12pt,english,compsoc,onecolumn]{IEEEtran}

\usepackage[T1]{fontenc}
\usepackage[latin1]{inputenc}
\usepackage{amsmath}
\usepackage{graphicx}
\usepackage{amssymb}
\usepackage{rotating}
\usepackage{url}
\usepackage{color}
\usepackage{array}
\usepackage{colortbl}
\usepackage{minibox}

\usepackage{multirow} % added by fei.
\usepackage[labelformat=simple]{subfig}
\newcolumntype{C}[1]{>{\centering\let\newline\\\arraybackslash\hspace{0pt}}m{#1}}

\newcommand{\comment}[1]{}

\newcommand{\argmax}{\operatornamewithlimits{arg\,max}}

\long\def\symbolfootnote[#1]#2{\begingroup\def\thefootnote{\fnsymbol{footnote}}
\footnote[#1]{#2}\endgroup}

\definecolor{orange}{cmyk}{0,0.5,0.8,0}
\definecolor{purple}{cmyk}{0.5,0.7,0,0}

\graphicspath{{Fig/}}

\begin{document}

\title{BreakingNews: Article Annotation by\\ Image and Text Processing}

\author{Arnau Ramisa*, Fei Yan*,  Francesc Moreno-Noguer,  \\and Krystian Mikolajczyk
\IEEEcompsocitemizethanks{
\IEEEcompsocthanksitem Arnau Ramisa and Francesc Moreno-Noguer are with the Institut de Rob\`otica i Inform\`atica Industrial, CSIC-UPC, Barcelona, 08028, Spain.  Email: \{aramisa, fmoreno\}@iri.upc.edu.
\IEEEcompsocthanksitem Fei Yan is with   the Centre for Vision, Speech and Signal Processing, University of Surrey, Guildford, UK. Email: f.yan@surrey.ac.uk.
\IEEEcompsocthanksitem Krystian Mikolajczyk is with the Department of Electrical and Electronic Engineering, Imperial College London, UK. Email: k.mikolajczyk@imperial.ac.uk. %
\newline \newline * The first two authors contributed equally.
\protect\\
}
}

\IEEEcompsoctitleabstractindextext{
\begin{abstract}
\boldmath
Building upon recent Deep Neural Network architectures, current approaches lying in the intersection of
computer vision and natural language processing have
achieved unprecedented breakthroughs in  tasks like automatic captioning or image retrieval. Most of these
learning methods, though, rely on large training sets of images associated with human annotations that specifically describe the visual
content. In this paper we propose to go a step further and explore the more complex cases
where textual descriptions are loosely related to the images. We focus on the particular domain of News articles in
which the textual content often expresses connotative and ambiguous relations that are only suggested but
not directly inferred from images.  We introduce new deep learning methods  that address  source detection, popularity prediction, article illustration and geolocation of articles. An adaptive CNN architecture is proposed, that shares most of the structure for all the tasks, and is suitable for multitask and transfer learning. Deep Canonical Correlation Analysis is deployed for article illustration, and a new loss function based on Great Circle Distance is proposed for geolocation.
Furthermore, we present BreakingNews, a novel dataset with approximately 100K news articles
including images, text and captions, and enriched with heterogeneous meta-data (such as GPS coordinates and popularity metrics). We show this dataset to be appropriate to explore all aforementioned problems, for which we provide a baseline performance using various Deep Learning architectures, and different
representations of the textual and visual features. We report very promising results  and bring to light
several limitations of current state-of-the-art in this kind of domain, which we hope will help spur progress in the field.
\end{abstract}

}

\maketitle

% % % % % % % % % % % % % % % % % % % % % % % % % % % % % % % % %
\section{Introduction}
\label{sec:Intro}
\IEEEPARstart{I}n recent years, there has been a growing interest in exploring the relation between images and language. Simultaneous progress in the fields of Computer Vision (CV) and Natural Language Processing (NLP) has led to impressive results in learning  both  image-to-text and text-to-image connections. Tasks such as automatic image captioning~\cite{ChenCVPR2015,DonahueCVPR2015,JohnsonICCV2015,karpathy15cvpr,VinyalsCVPR2015,XuICML2015}, image retrieval~\cite{FarhadiECCV2010,HodoshJAIR2013,KovashkaIJCV2015,KumarECCV2008} or image generation from sentences~\cite{ChangSPONSOR2014,ZitnickICCV2013}  have shown unprecedented results, which claimed to be similar to the performance  expected from a three-year old child\footnote{\url{http://www.ted.com/talks/fei_fei_li_how_we_re_teaching_computers_to_understand_pictures}}.

\begin{figure*}[t!]
\begin{center}
\includegraphics[width=0.48\textwidth]{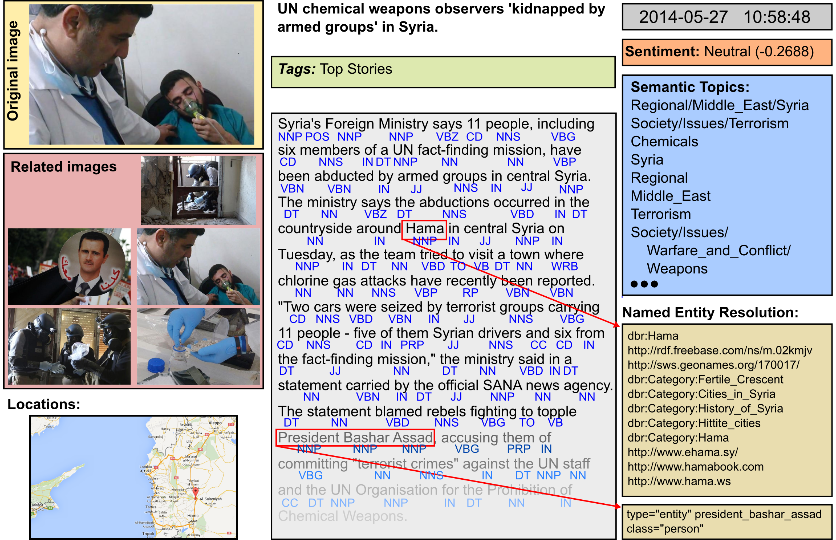}
\includegraphics[width=0.48\textwidth]{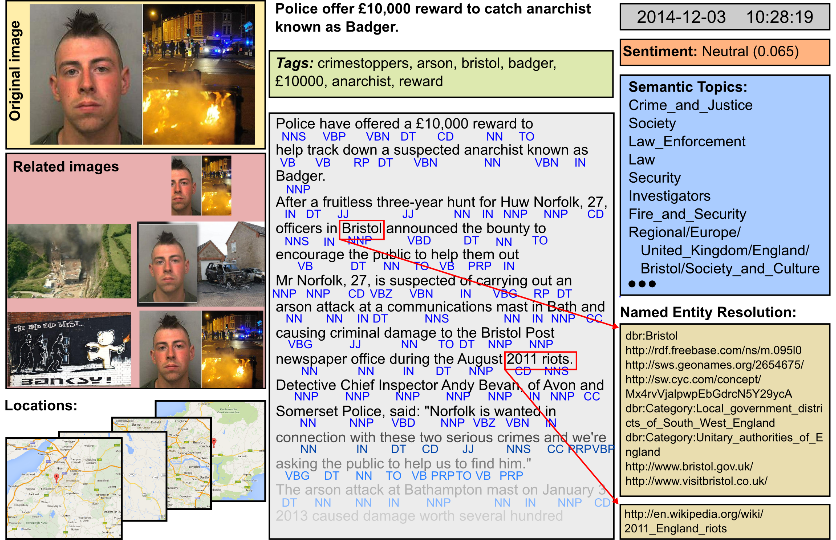}
\end{center}
\caption{{\bf BreakingNews dataset}. The dataset contains a variety of news-related information including: the text of the article, captions, related images, part-of-speech tagging, GPS coordinates, semantic topics list or results of sentiment analysis, for about 100K news articles. The figure shows to samples images. All this volume of heterogeneous data makes   BreakingNews an appropriate  benchmark for several tasks exploring the relation between text and images. In this paper we show a few of them. }
\label{fig:intro}
\end{figure*}

One of the main reasons behind the  success of these approaches is the resurgence of deep learning for modeling data, which has been possible  due to the development of new  parallel computers and GPU  architectures and  due to the release of new large datasets, used to train deep models with many parameters.

The popularity of crowd sourcing tools has facilitated the proliferation of a number of  these datasets combining visual and language content. Among them, the most widely known are the UIUC Pascal Sentence Dataset~\cite{RashtchianNAACL2010}, the SBU captioned photo dataset~\cite{OrdonezNIPS2011}, Flickr8K~\cite{HodoshJAIR2013}, Flickr30K~\cite{YoungTACL2014} and MS-COCO~\cite{LinECCV2014}. All these datasets consist of a number of images (from 1K to 1M), each annotated by  human written sentences (between 1 and 5 per image). The annotations are typically short and accurate sentences (of less than 20 words) describing the visual content of the image and the action taking place. In contrast, other and more complex types of documents, like illustrated news articles, have been barely explored.

We believe that current successes in the crossroads between NLP and computer vision indicate that the techniques are mature for more challenging objectives than those posed by existing datasets. The NLP community has been addressing tasks such as sentiment analysis, popularity prediction, summarization, source identification or geolocation  to name a few that have been relatively little explored in computer vision. In this paper we propose several learning schemes. Specifically, for the source detection, article illustration, popularity  and geolocation prediction, we consider an adaptive CNN architecture, that shares most of the structure for all the problems, and just requires replacing and retraining the last layers in order to tackle each particular problem. We also compare single-task, multi-task and transfer learning on mixed NLP/CV tasks. For the caption generation task, image and text representations are combined in a Long Short-Term Network (LSTM).  

In order to evaluate these algorithms,  we have collected {\em BreakingNews}, a large-scale dataset of news articles with rich meta-data. Our dataset consists of approximately 100K news articles, illustrated by one to three images and their corresponding captions. Additionally, each article is  enriched with other data like related images from Google Images, tags, shallow and deep linguistic features (e.g.  parts of speech, semantic topics or outcome of a sentiment analyzer), GPS latitude/longitude coordinates and reader comments. The articles cover the whole year 2014 and are collected from various reference newspapers and media agencies like BBC News, The Guardian or the Washington Post. 

This dataset is an excellent benchmark for taking joint vision and language developments a step further. In contrast to existing datasets, the link between images and text in BreakingNews  is not as direct, i.e., the  objects, actions and attributes of the images may not explicitly appear as words in the text (see examples in Fig.~\ref{fig:intro}). The visual-language connections are more subtle and learning them will require the development of new inference tools able to reason at a higher and more abstract level. Furthermore, besides tackling article illustration or image captioning  tasks, the proposed dataset is intended to address new challenges, such as source/media agency detection, estimation  of GPS coordinates,  or popularity prediction  (which we annotate based on  the reader comments and number of re-tweets).

For each of these tasks we benchmark several learning schemes based on  deep neural network architectures  and different feature representations for  text and images. Overall results are very promising, but there is still much room for improvement, and the dataset can be easily extended with additional annotations to explore other problems such as sentiment analysis or text summarization. Both the baseline results we obtain and the dataset will be made publicly available, and we hope it will inspire future research in the field.

\vspace{2mm}
\noindent{\bf Overview:} The rest of the paper is organized as follows. In Section~\ref{sec:related} we present the different vision and language processing tasks that will be addressed in this paper, with the review of the related work for each case. We consider this thorough  revision of  related work   in up to seven different text-and-image topics,  to be one of the main contributions of the paper. The following two sections, describe the technical details about how the visual and textual data is represented (Section~\ref{sec:repr}), and which CNN architectures we have built to tackle each of the tasks (Section~\ref{sec:learn}). In Section~\ref{sec:dataset} we introduce {\em BreakingNews}, the news article dataset that we will use to evaluate our CNNs for each of the tasks, in Section~\ref{sec:results}.

% % % % % % % % % % % % % % % % % % % % % % % % % % % % % % % % %

\section{Tasks Description and Related Work}
\label{sec:related}

We next describe the tasks that will be tackled in this article and the related work for each of them, as well as the existing datasets in Section~\ref{sec:dataset}.

\subsection{Text Illustration}
\label{sec:RelatedIllustration}

This task deals with automatically retrieving the appropriate image (or a small subset) from a pool of images given a textual query of a news story. 

Some approaches  tackle this problem by   learning classifiers to represent images through intermediate semantic concepts, that can then be  easily assigned to individual keywords or to multi-attribute textual descriptions~\cite{BarnardJMLR2003, BarnardICCV2001,KovashkaIJCV2015,KumarECCV2008,RasiwasiaToM2007,DouzeCVPR2011}. Richer textual queries are allowed in~\cite{FarhadiECCV2010,HodoshJAIR2013} by mapping both image and sentences to intermediate spaces where  direct comparison is possible.  For all these methods, the textual input consists on keywords or short sentences at most.

There have been some efforts specifically addressing the domain of news articles. For instance~\cite{FengACL2010} builds a system based on joint topic models to link text to images, and evaluates the results in the BBC News dataset~\cite{FengPAMI2013}. In~\cite{CoelhoLNCS2012}, it is assumed that  there exist short text descriptions and tags accompanying the images, which can then be easily matched to text documents represented by means of word frequencies. Other approaches perform article illustration by exploiting Google's search engine (which also assumes text associated with each image of the database), and combine multiple queries generated from the title of the article~\cite{LiICM2011}, or from the narrative keywords~\cite{HuangCTAA2013}. Similarly,~\cite{JoshiTOMCCAP2006} proposes a story picturing engine, that first processes a story to detect certain keywords, and then uses these to select annotated images from a database. 

Alternatively to image retrieval strategies, text illustration can be carried out from a generative perspective. There exist early approaches that extracted object arrangements from sentences to then generate  computer graphic representations of static~\cite{CoyneCCGIT2001} and dynamic~\cite{PerlinCCGIT1996} scenes.~\cite{GobronVC2010} proposed a system to animate a human avatar based on the emotions inferred from text. And very recently, advanced sentence parsers have been used to extract objects and their relations from sentences,  and then automatically render 2D~\cite{ZitnickICCV2013} and 3D~\cite{ChangSPONSOR2014} scenes. 

Finally, illustration of short texts, like chat messages or tweets, has also been investigated~\cite{JiangMS2014,WangACM2014}.

\subsection{Geolocation}
\label{sec:RelatedGeolocation}
This task considers the problem of geographically referencing news articles based on text and image cues. One of the pioneering works on a related topic proposed an approach for geolocating web documents  by detecting and disambiguating location names in the text~\cite{DingICVLDB2000}. Also only using text information,~\cite{SerdyukovCRDIR2009} matched a predefined tagged map to the most likely  GPS location of tagged Flickr photos. On the other hand,  several image-based methods leverage massive datasets  of geotagged images for geolocation of generic scenes on the global Earth scale~\cite{HaysCVPR2008,KalogerakisICCV2009,WeyandARXIV2016}, or for place recognition tasks~\cite{ChenCVPR2011}. There has been also work on this area by combining text and image information~\cite{CaoICM2009,CrandallICWWW2009}.

When addressing the specific domain of news articles, most existing works use only text information~\cite{LiICM2011,WingACL2011}. One interesting exception is~\cite{ZhouICM2012}, which combines textual descriptors with global image representations based on GIST~\cite{OlivaPBR2006} to infer geolocation of nearly two thousand NY Times articles. 

\subsection{Popularity Prediction}
\label{sec:PopPredict}

Predicting the  popularity of a blog post, an article, a video, or even a tweet (i.e.\ how many times it has been viewed or shared, or how many comments it generated) has useful applications for online publishers and social networks, and consequently has attracted the interest of many researchers~\cite{BandariARXIV2012,BaoICWWW2013,PintoICWSD2013, TatarSNAM2014,TsagkiasAIR2010}. This is also related to recent approaches that address the question of what makes an image memorable, interesting or popular~\cite{DharCVPR2011,GygliICCV2013,IsolaPAMI2014}.

In the case of news articles, the number comments an article can generate~\cite{TatarSNAM2014, TsagkiasAIR2010}, or the number of times it has been shared on social networks~\cite{BandariARXIV2012} is often used as a proxy for its popularity. This measures are then used to predict which articles will be more relevant, or its average ``life span''.

\subsection{Source detection}
\label{sec:RelatedSource}
This tasks deals with analyzing the content of the news articles, and
detecting in which news media agency it has been originally
published. This task can be resolved by modeling the type of language and
topical preference for each news agency, but also from correctly
modeling the sentiment in each news article based on the political
orientation of the agency, or other similar refinements. 

Although we are not aware of other approaches explicitly tackling source identification in news articles, there exists a vast amount of related works, mostly motivated by detection of plagiarism or  dealing with the authorship identification problem~\cite{KoppelLRE2011,NarayananSP2012,StamatatosJASIS2009}.

\subsection{Caption Generation}
\label{sec:RelatedCapGen}

Among the tasks dealing with image and text interactions, automatically generating photo captions is the one  receiving most attention.  Early work in this area focused in annotating images with single words~\cite{BarnardJMLR2003,SocherCVPR2010}, phrases with a reduced vocabulary~\cite{FarhadiECCV2010,KulkarniCVPR2011} or with semantic tuples~\cite{OrdonezIJCV2015,QuattoniNAACL2016}. The problem has also been tackled from a ranking perspective, in which given a query image, one needs to rank a set of human generated captions. Projecting images and captions to a joint representation space using Kernel or Normalized Canonical Correlation Analysis   has been used for this purpose~\cite{GongECCV2014,HodoshJAIR2013}. The largest body of recent work, though,  share a basic approach, which consists in using a  Recurrent Neural Network (RNN) to learn to ``translate'' the image features into a sentence in English, one word at a time~\cite{ChenCVPR2015,DonahueCVPR2015,FangCVPR2015,JohnsonICCV2015,karpathy15cvpr,KirosTACL2015,MaoICLR2015,VinyalsCVPR2015,XuICML2015}. 

Yet, while the results obtained by the RNN-based approaches are outstanding, they are all focused on accurately describing the content of the picture, which is significantly different than generating a suitable caption for the illustration of a news article,  in which the relation between the visual content of the image and the corresponding captions is more indirect and subtle than in typical image captioning datasets. This is precisely the main challenge posed by the BreakingNews dataset we present. There exists some previous work in automatic caption generation for images of the BBC News Dataset~\cite{FengPAMI2013}. However, as we will discuss later, BreakingNews is about two orders of magnitude larger than BBC News Dataset, and incorporates a  variety of metadata that brings the opportunity to develop  new computer vision approaches for the analysis of  multi-modal and large-scale multimedia data.

\subsection{Image and Text Datasets}
\label{sec:RelatedDatasets}
There is no doubt that one of the pillars on which the recent advent of Deep Learning holds is the proliferation of large object classification and detection datasets like 
ImageNet~\cite{DengCVPR2009}, PASCAL VOC 2012~\cite{EveringhamIJCV2010} and SUN~\cite{XiaoCVPR2010}.
Similarly, the progress on the joint processing of natural language  and images largely depends on several  datasets of images  with human written descriptions. For instance, the UIUC Pascal Sentence Dataset~\cite{RashtchianNAACL2010} consists of 1,000 images each annotated by 5 relatively simple sentences, even lacking verb in a large percentage of them. The SBU photo dataset~\cite{OrdonezNIPS2011} consists of one million web images with one description per image. These descriptions are automatically mined and do not always describe the visual content of the image. The Flickr8K~\cite{HodoshJAIR2013}, Flickr30K~\cite{YoungTACL2014} and MS-COCO~\cite{LinECCV2014} contain five sentences for a collection of 8K, 30K and 100K images, respectively. In these datasets,  the  sentences specifically describe the visual content and actions occurring in the image. 

In this paper, we focus on illustrated news articles datasets, which differ from  previous image and text datasets in that  the captions or the text of the article do not necessarily describe the objects and actions in the images.  The relation between text and images may express connotative relations rather than specific content relations. The only dataset similar to BreakingNews is the BBC News dataset~\cite{FengPAMI2013}, which contains up to 3,000 articles with images, captions and body text.  However, this dataset is mainly intended for caption generation tasks. In BreakingNews, we provide a two order of magnitude larger dataset, that besides images and text, contains other metadata (GPS location, user comments, semantic topics, etc) which allows exploring a much wider set of problems and exploit the power of current Deep Learning algorithms.

\subsection{Multi-Task and Transfer Learning on Large Datasets } As just said, one of the distinguishing features of BreakingNews   is that it is a large-scale  multimodal and heteregeous dataset (see Fig.~\ref{fig:intro}),  appropriate for  exploring different and learning algorithms. These  algorithms can, in turn,  be independently developed for each of the tasks, or as we will show, we can also explore methodologies  for multi-task  and transfer learning, which as demonstrated in~\cite{LongARXIV2015,YosinskiNIPS2014} can produce a boost to generalization of  modern deep learning networks.

% % % % % % % % % % % % % % % % % % % % % % % % % % % % % % % % %
\section{Representation}
\label{sec:repr}

In this section we discuss text and image representations for news article analysis. We first present Bag-of-Words and Word2Vec text embeddings and then image representations based on deep CNNs.

\subsection{Text representation}
\label{subsec:repr:text}

{\noindent \bf Bag-of-Words (BoW) with TF-IDF weighting: } 
Bag-of-Words is one of the most established text representations, and their reliability and generalization capabilities have been demonstrated in countless publications. Therefore, we adopt it as a baseline for our approach. The BoW representation requires a vocabulary, which we established as unique lemmatised tokens from the training data that appear more than $L$ times in the articles. This leads to $D_b$ dimensional Bag-of-Words (BoW) vectors where the $j^{\textrm{th}}$ dimension is given by $t^j \log \frac{M}{c^j+1}$, where $t^j$ is the term frequency of the $j^{\textrm{th}}$ token (i.e. the number of times it appears in the article), $c^j$ is document frequency of the token (i.e. the number of training articles where it appears), and $M$ is the total number of training articles. A common practice is to truncate the BoW vector based on inverse document frequency. It has been demonstrated in various studies that the performance in various tasks improves monotonically as the number of retained dimensions increases.

\vspace{2mm}
{\noindent \bf Word2Vec: } Recently, distributed representations for text, and \emph{Word2vec}~\cite{le14icml,mikolov13iclr,mikolov13nips} in particular, are gaining a lot of attention in the NLP community. This representation encodes every word in a real-valued vector that preserves semantic similarity, e.g. "king" minus "man" plus "woman" will be close to "queen" in the embedding space. Using a two-layer neural network, words are modeled based on their context, defined as a window that spans both past and future words. Two methods have been proposed to learn the representations: the \emph{Continuous bag of words} (cbow) model, where the objective is predicting a word given its context, and the \emph{Continuous skip-gram} model, where the objective is the opposite; i.e.~trying to predict the context given the word being modeled. The \textit{negative sampling} objective function, where the target word has to be distinguished from random negative words, is used as an efficient alternative to hierarchical soft max. We investigated both cbow and skip-gram methods\footnote{Code available at https://code.google.com/p/word2vec/} and found that the later performed better in our applications.

The trained model is used to encode each article by stacking the representation for every word and symbol in a matrix with dimension $D_w$ times the number of tokens in the article, where $D_w$ is the size of the embedding space. From this matrix we investigate three representations: 1) full matrix as input for the convolutional layers of a deep network; 2) mean or 3) max of each dimension to construct a single $s_e$ dimensional vector for each article.

\begin{figure*}[t!]
\begin{center}
\includegraphics[height=8cm]{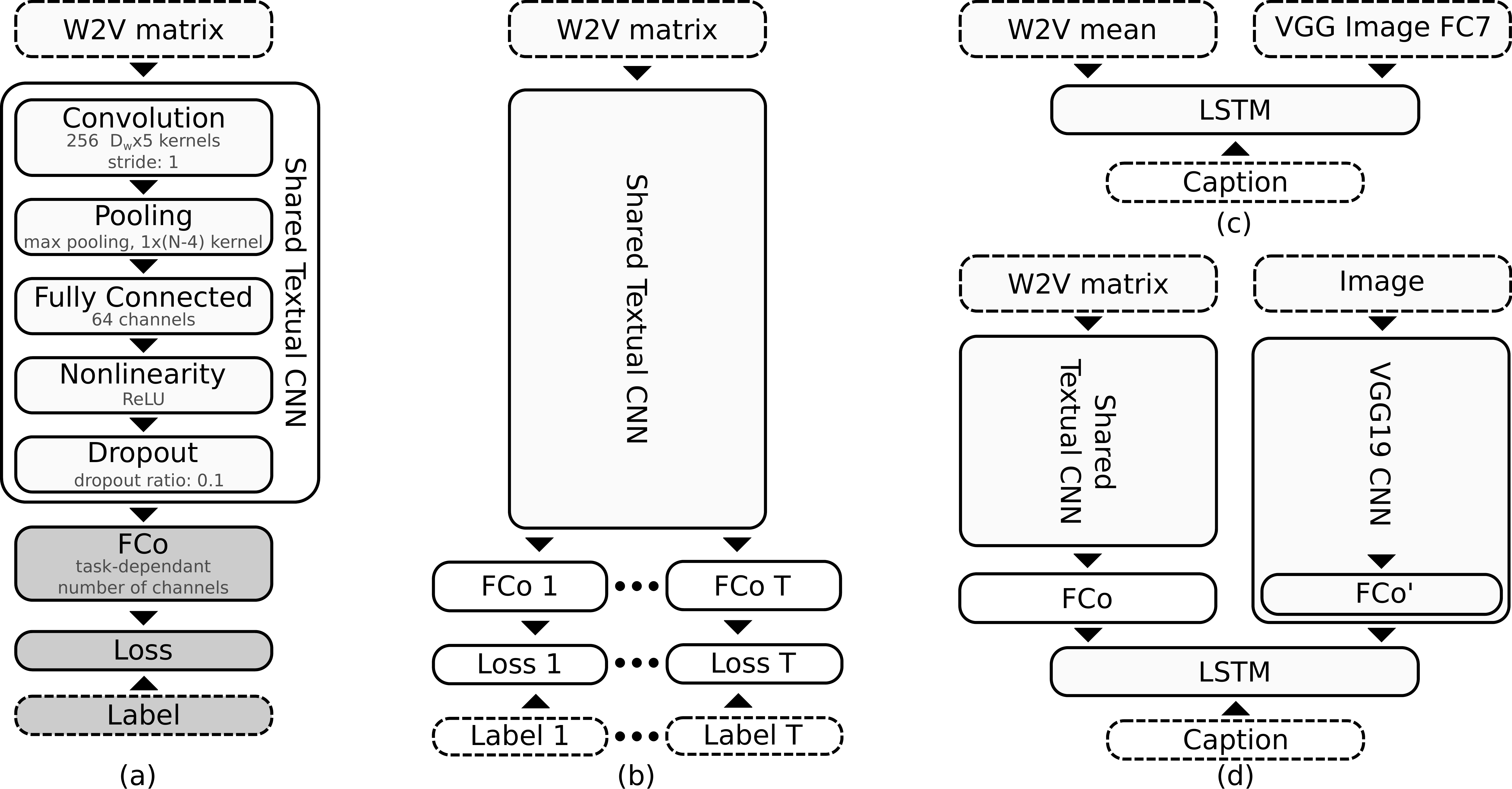}
\end{center}
\caption{{\bf CNN architectures for article analysis and caption generation.} (a) CNN for article analysis, used in the tasks of article illustration,  source detection and popularity and geolocation prediction. Dashed boxes are the CNN inputs, and solid boxes correspond to the layers. White boxes (enclosed within the "Shared Textual CNN" box) are shared by all tasks, and shaded boxes are task specific layers. (b) Multitask CNN for T tasks. (c, d)  Models for caption generation: (c) LSTM with fixed features, and (d) end-to-end learning with CNNs+LSTM. Note that the "Shared Textual CNN" from (a) is used in the textual branch.}\label{fig:cnn_arch}
\end{figure*}

\subsection{Image representation}
\label{subsec:repr:image}

Our image representations are based on pre-trained deep CNNs. We follow the state-of-the-art CNN representation developed for object recognition~\cite{simonyan14arxiv} with 19 convolutional layers (VGG19). More specifically, we compute the activations of the last ReLU layers of the VGG19  and the Places scene recognition CNN~\cite{zhou14nips}. Each of these activations are 4,096 dimensional sparse vectors, and have been shown to perform well in various vision tasks. We also $\ell_2$ normalize the two vectors and concatenate them to form a third image representation of size 8,192.

% % % % % % % % % % % % % % % % % % % % % % % % % % % % % % % % %
\section{Learning}
\label{sec:learn}

In this section we describe the learning schemes based on deep neural networks that will be used for the proposed tasks. We first discuss in~\ref{subsec:learn:cnn} a CNN for article illustration, source detection, and popularity and geolocation prediction. In~\ref{subsec:learn:lstm} we extend the state-of-the-art LSTM approach for caption generation.

\subsection{CNN for article analysis}
\label{subsec:learn:cnn}

{\noindent \bf CNN architecture: }CNNs have recently proven successful in many NLP tasks such as sentiment analysis, machine translation or paraphrase identification~\cite{collobert11jmlr,kalchbrenner14acl,kim14emnlp}. We propose a CNN architecture, as illustrated in Fig.~\ref{fig:cnn_arch}a, for various article analysis tasks. In the figure, dashed boxes are the inputs to the network, and solid boxes are the layers. Moreover, white boxes denote data and layers that are shared by all tasks (we denote the group by "Shared Textual CNN"), while shaded ones are task specific.

Assuming Word2Vec embeddings have been learnt, an article can  be represented by a $D_w \times n_i$ matrix, where $D_w$ is the dimensionality of the embedding space, and $n_i$ is the number of tokens in the $i^{\textrm{th}}$ article. Such matrix is zeros-padded to $D_w\times N$, where $N$ is  the number of tokens in the longest article (we add $N-n_i$ columns of zeros). The \emph{convolution} layer convolves the Word2Vec matrices with 256 rectangular kernels of size $D_w \times 5$ at a stride of 1, capturing local correlations in the tokens. The output of the convolution layer are $256$ feature channels of dimension $N-4$ each. Effectively, the convolutional kernels act as learnable feature detectors. 

The activations of the convolution layer are max-pooled (\emph{pooling}) along each channel. The resulting 256 dimensional vectors are transformed into 64 channels in a fully connected linear layer (\emph{FC}), before \emph{nonlinearity} is applied. Although Tanh or Absolute nonlinearities were observed to have a small accuracy advantage, we found that ReLU~\cite{krizhevsky12nips} is numerically more stable in terms of gradient computation.

\emph{Dropout} has been proven a simple yet effective way of preventing overfitting in large networks~\cite{srivastava14jmlr}. We found that a small dropout ratio (e.g. 0.1) is sufficient for our tasks. Finally, after dropout, the 64 dimensional vectors are transformed in a second linear layer (\emph{FCo}) to vectors with appropriate length. Let the output of \emph{FCo} be ${\bf z} \in \mathbb{R}^d$ for each article, which will then be used in a task specific loss function, together with a task specific label. The gradient of the loss with respect to $\bf z$ will be computed and backpropagated to update the CNN parameters. The parameters of the CNN architecture for various tasks are given in Fig.~\ref{fig:cnn_arch}a. 

We next describe the  configuration details of the last layer \emph{FCo} and the loss functions (and when necessary their corresponding derivatives), that make the CNN architecture of Fig.~\ref{fig:cnn_arch}a problem specific.

\vspace{2mm}
{\noindent \bf Source detection: } For source detection, the label $y$ is in the set $\{1,\cdots,d\}$, being $d$ the number of  news agencies where the articles originate from. The \emph{loss} layer implements the multinomial logistic loss that maximizes the cross entropy, whose gradients for backpropagation are readily available.

\vspace{2mm}
{\noindent \bf Article illustration: } For article illustration, we use the image representation discussed in Sec.~\ref{subsec:repr:image} as the label ${\bf y} \in \mathbb{R}^{8,192}$, which is a concatenation of $\ell_2$ normalized VGG19 and Places activations. The output of the \emph{FCo} layer $\bf z$ also has a dimensionality $d=8,192$. For a batch of $m$ data points, the inputs to the \emph{loss} layer are two $d \times m$ matrices $Z$ and $Y$.

Let the covariances of $Z$ and $Y$ be $\Sigma_{zz}$ and $\Sigma_{yy}$ respectively, and let the cross covariance be $\Sigma_{zy}$, all of them $\mathbb{R}^{d\times d}$ matrices. A good loss function for article illustration is the Canonical Correlation Analysis (CCA) loss, which seeks pairs of linear projections ${\bf w}_z,{\bf w}_y \in \mathbb{R}^d$ that maximize the correlation between the articles and the images:
\begin{eqnarray}
({\bf w}_z^*, {\bf w}_y^*) &=& \argmax_{{\bf w}_z,{\bf w}_y} \hspace{1mm} \textrm{corr}({\bf w}_z^\top Z, {\bf w}_y^\top Y) \nonumber \\
&=& \argmax_{{\bf w}_z,{\bf w}_y} \frac {{\bf w}_z^\top \Sigma_{zy} {\bf w}_y} {\sqrt{{\bf w}_z^\top \Sigma_{zz} {\bf w}_z {\bf w}_y^\top \Sigma_{yy} {\bf w}_y}}
\label{eq:cca_single}
\end{eqnarray}
Using the fact that the objective is invariant to scaling of ${\bf w}_x$ and ${\bf w}_y$, and assembling the top projection vectors into the columns of projection matrices $W_z$ and $W_y$, the CCA objective can be written as:
\begin{align}
\label{eq:cca_all}
& \hspace{5mm} \max_{W_z,W_y} \hspace{1mm} \textrm{tr}(W_z^\top \Sigma_{zy} W_y) & \\
& \textrm{s.t.}: W_z^\top \Sigma_{zz} W_z = W_y^\top \Sigma_{yy} W_y = I_d & \nonumber
\end{align}
where $I_d$ is the $d-$dimensional identity matrix.

Let us define $\Gamma=\Sigma_{zz}^{-1/2} \Sigma_{zy} \Sigma_{yy}^{-1/2}$, and let $U_k$ and $V_k$ be the matrices of the first $k$ left- and right- singular vectors of $\Gamma$,  respectively. In~\cite{andrew13icml,mardia79}  it is shown that the optimal objective value is the sum of the top $k$ singular values of $\Gamma$, and the optimum is attained at
\begin{equation}
(W_z^*, W_y^*) = (\Sigma_{zz}^{-1/2}U_k,\Sigma_{yy}^{-1/2}V_k)
\label{eq:optimum_attained_at}
\end{equation}
When $k=d$, the total correlation objective in Eq.~\eqref{eq:cca_all} is equal to the trace norm of $\Gamma$:
\begin{equation}
\textrm{corr}(Z,Y) = ||\Gamma||_{\textrm{tr}} = \textrm{tr}((\Gamma^\top \Gamma)^{1/2})
\label{eq:trace_norm}
\end{equation}

We define the CCA-based loss as $L=-\textrm{corr}(Z,Y)$. If we consider the singular value decomposition (SVD) of $\Gamma$ to be $\Gamma = UDV^{\top}$, recent works~\cite{andrew13icml,yan15cvpr}, have shown that  the gradient of $L$ with respect to $Z$ is given by:
\begin{equation}
\frac {\partial L} {\partial Z} = -\frac{1}{m-1} (2 \nabla_{zz}{\bar Z} + \nabla_{zy}{\bar Y})
\label{eq:gradient_for_z}
\end{equation}
where $\bar Z$ and $\bar Y$ are the centered data matrices, and
\begin{eqnarray}
& \nabla_{zz} = -\frac{1}{2} \Sigma_{zz}^{-1/2} UDU^\top \Sigma_{zz}^{-1/2} & \\
& \nabla_{zy} = \Sigma_{zz}^{-1/2} UV^\top \Sigma_{yy}^{-1/2} &
\label{eq:nabla_for_z}
\end{eqnarray}

\begin{figure}[t!]
\centering
\vspace{-6mm}
\includegraphics[width=85mm]{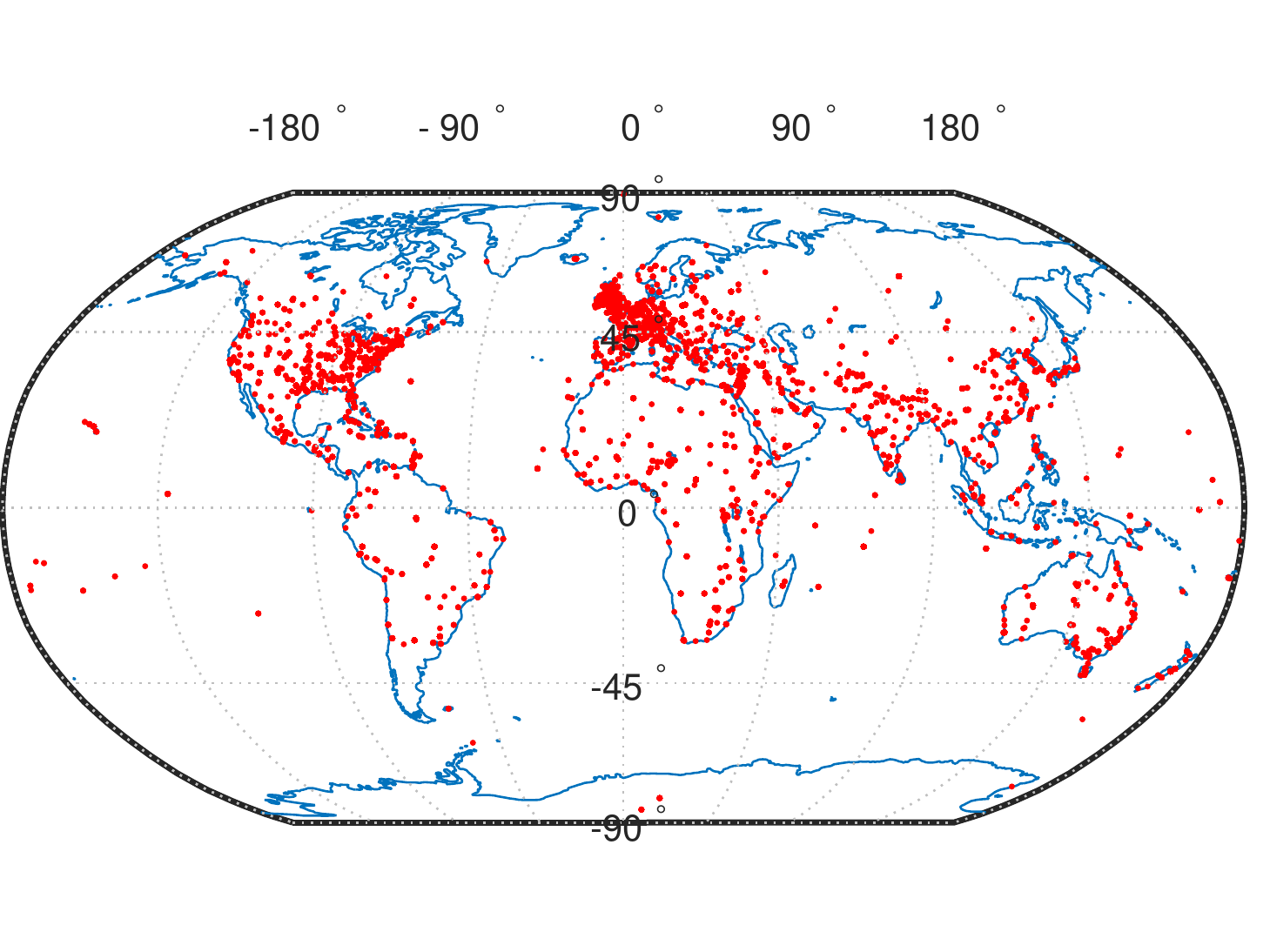}
\vspace{-6mm}
\caption{Ground truth geolocations of the articles in training set.}
\label{fig:geoloc_gt}
\end{figure}

\vspace{2mm}
{\noindent \bf Popularity prediction: } For this task the label $y$ is a scalar denoting the number of comments. Accordingly the output of the \emph{FCo} layer $z$ has dimension $d=1$. In the \emph{loss} layer we minimize the $\ell_1$ distance between $z$ and $y$, and backpropagate the gradient.

\vspace{2mm}
{\noindent \bf Geolocation prediction: } For geolocation prediction, the label ${\bf y} =[y_1,y_2]^\top$ is a pair of latitude and longitude values, where the latitude $y_1 \in [-\pi/2, \pi/2]$ and the longitude $y_2 \in [-\pi, \pi]$. Fig.~\ref{fig:geoloc_gt} illustrates the ground truth geolocations in the training set.

The output of the \emph{FCo} layer ${\bf z} \in \mathbb{R}^2$ can be thought of as the predicted latitude and longitude. In the \emph{loss} layer, we could minimize the Euclidean distance between the two geolocations $\bf z$ and $\bf y$. However, the Euclidean loss does not take into account the fact that the two meridians near $-\pi$ and $\pi$ respectively are actually close to each other. To address this, we minimize the Great Circle Distance (GCD) between $\bf z$ and $\bf y$, which is defined as the geodesic distance on the surface of a sphere, measured along the surface. We use the spherical law of cosines approximation of GCD:
\begin{equation}
\label{gcd}
\textrm{GCD} = R \cdot \arccos(\sin y_1 \sin z_1 + \cos y_1 \cos z_1 \cos\delta)
\end{equation}
where $\delta = z_2 - y_2$, and $R=6,137$ km is the radius of the Earth. Ignoring the constant $R$ we define our geolocation loss function as
\begin{equation}
\label{loss_gcd}
L = \arccos(\sin y_1 \sin z_1 + \cos y_1 \cos z_1 \cos\delta)
\end{equation}
Using the chain rule, the gradient of $L$ with respect to $\bf z$ can be shown to be:
\begin{eqnarray}
\frac{\partial L}{\partial {\bf z}} = 
\left( \begin{array}{ll}
-\frac{1}{\sqrt{1-\phi^2}} ( \sin y_1 \cos z_1 - \cos y_1 \sin z_1 \cos\delta) \\
-\frac{1}{\sqrt{1-\phi^2}} (-\cos y_1 \cos z_1 \sin\delta)
\end{array} \right) 
\label{geo_grad}
\end{eqnarray}
where $\phi = \sin y_1 \sin z_1 + \cos y_1 \cos z_1 \cos\delta$.  In practice, to ensure numerical stability,  the term $1-\phi^2$ is discarded when it is too close to zero.

\begin{figure*}[t!]
\captionsetup[subfigure]{labelformat=empty}
\centering
\includegraphics[width=32mm]{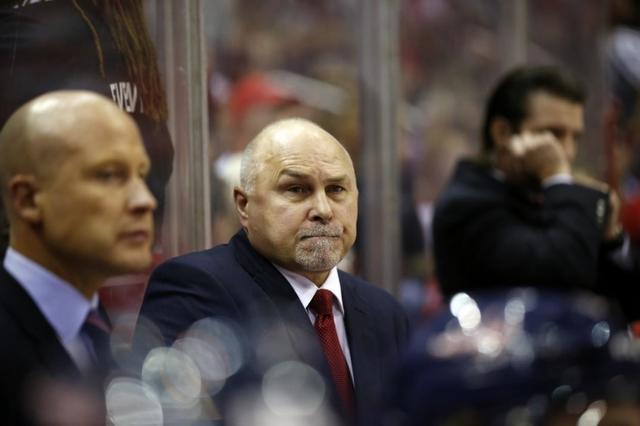} \hspace{0.5mm} \parbox[b][18mm][c]{4cm}{\small (AP Photo/Alex Brandon)} \hspace{4mm} \includegraphics[width=32mm]{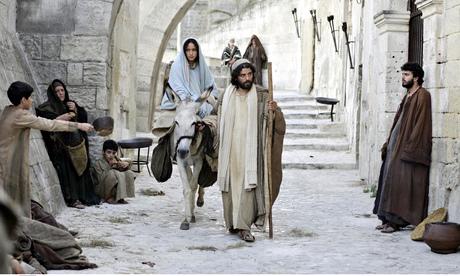} \hspace{0.5mm} \parbox[b][18mm][t]{6cm}{\small Despite long distances and late pregnancy, Mary and Joseph made an effort to pay their taxes. Photograph: Rex Features} \\ \vspace{0.3mm}
\includegraphics[width=32mm]{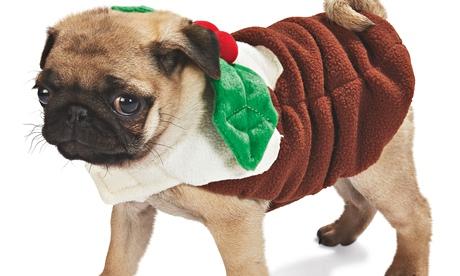} \hspace{0.5mm} \parbox[b][18mm][c]{4cm}{\small A dog wearing a Pets at Home outfit. Photograph: Tim Ainsworth} \hspace{4mm} \includegraphics[width=32mm, clip=true, trim=0 0 0 25mm]{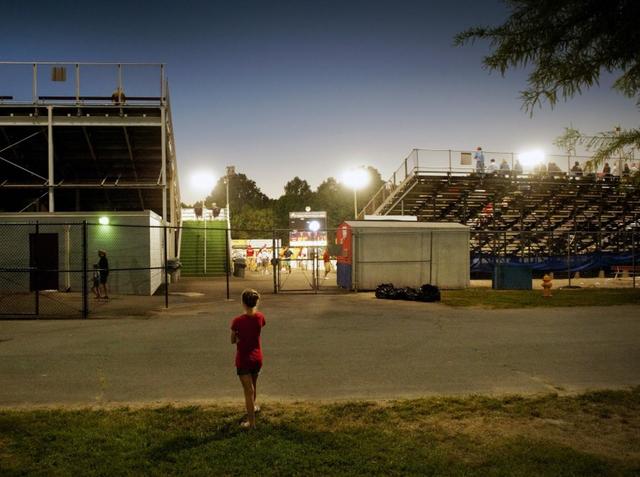} \hspace{0.5mm} \parbox[b][18mm][t]{6cm}{\small A young fan waits outside the stadium where the minor league team the P-Nats play in Dale City, Va. (Bill O'Leary/The Washington Post)} \\ \vspace{0.3mm}
\includegraphics[width=32mm]{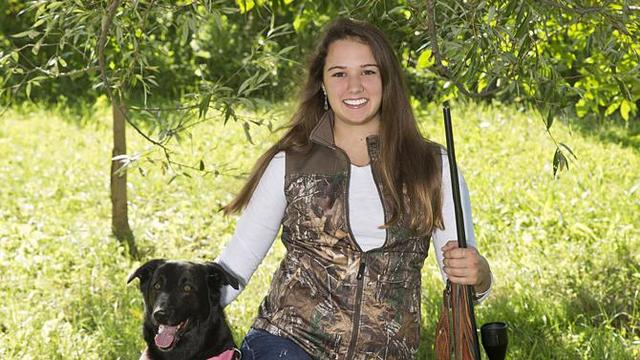} \hspace{0.5mm} \parbox[b][18mm][c]{4cm}{\small Rebekah Rorick yearbook photo} \hspace{4mm} \includegraphics[width=32mm]{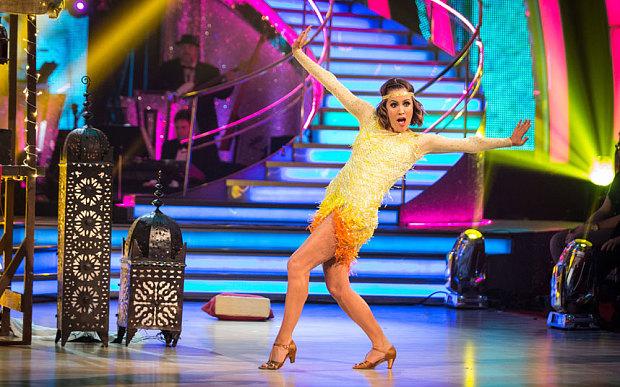} \hspace{0.5mm} \parbox[b][18mm][c]{6cm}{\small Caroline Flack won the Strictly Come Dancing finale, but could be a turn-off on stage, research suggests}\\
\vspace{1mm}
\caption{Example images and  their corresponding captions in the training set. Note that  the captions not always  explicitly describe the visual content of the images.} % \fei{don't know how to do the 3x2 version.. can someone else give it a go?}}
\label{fig:example_captions}
\end{figure*}

\vspace{2mm}
{\noindent \bf Multitask and transfer learning: } Multitask   and transfer learning are both effective methods for sharing learnt information. Information in deep neural networks is learnt in the form of hierarchical representations, which allows layer specific sharing and transfer of parameters. We  consider the following two variants of the CNN:
\begin{itemize}
  \item Multitask CNN: At the output of the CNN we have multiple parallel loss layers and the corresponding \emph{FCo} layers (see Fig.~\ref{fig:cnn_arch}b). The goal is to learn representations shared by multiple tasks in the layers of the "Shared Textual CNN" block, and task-specific representations in the \emph{FCo} layers.
  \item Transfer CNN: A CNN is first trained in a single task setting. The loss layer and corresponding \emph{FCo} layer are then replaced with those for a new task, with the parameters for these layers randomly initialized, and the model is finetuned on the new task.
\end{itemize}

\subsection{LSTM for caption generation}
\label{subsec:learn:lstm}

Image and video captioning has made significant progress in the last few years. State-of-the-art approaches use CNNs to encode image and video as a fixed-length vectors, and employ recurrent neural networks (RNNs) in the particular form of Long Short-Term Memory Networks (LSTMs)~\cite{hochreiter97nc} to decode the vector into a caption~\cite{DonahueCVPR2015,VinyalsCVPR2015,XuICML2015}. By using an input gate, an output gate and a forget gate, LSTMs are capable of learning long-term temporal dependences between important events. Moreover, LSTMs successfully deal with the vanishing and exploding gradients problem by keeping the errors being backpropagated in their memory.

In this paper we  consider the following LSTM-based architectures to tackle the caption generation problem. 

\vspace{2mm}
{\noindent \bf LSTM with fixed features: } The task of news image captioning differs from most existing works on image captioning in two aspects. First, captions for news images are often only loosely related to what is illustrated in the images, making it much more challenging to learn the mapping between images and captions. Six example images and associated captions in the training set are shown as examples in Fig.~\ref{fig:example_captions}. Second, instead of conditioning only on images, news captions are conditioned both on the articles and images. To address the second difference, we extend the recent techniques in~\cite{DonahueCVPR2015,VinyalsCVPR2015,XuICML2015}. We take the mean of a Word2Vec matrix as the fixed article representation, and VGG19 and/or Places activations as the fixed image representation, and train an LSTM using either representation, or a concatenation of the two. The architecture of such a scheme is illustrated in Fig.~\ref{fig:cnn_arch}c.

\vspace{2mm}
{\noindent \bf End-to-end learning with CNNs+LSTM: } The language CNN described earlier in this section (Fig.~\ref{fig:cnn_arch}a), the VGG19 (or Places) image CNN, and the LSTM for caption generation are all neural networks trained with backpropagation. We propose to connect the three networks as illustrated in Fig.~\ref{fig:cnn_arch}d, achieving end-to-end learning of caption generation from raw articles and raw images. In this setting, the VGG19 CNN and the language CNN have \emph{FCo'} and \emph{FCo} as their last layers respectively, which encode image and article information into a 500 dimensional vector each. The VGG19 CNN is also filled with parameters pretrained on ImageNet. During training, the gradients in LSTM are propagated backward in time, and backward into the two CNNs.

% % % % % % % % % % % % % % % % % % % % % % % % % % % % % % % % % %
\section{BreakingNews Dataset}
\label{sec:dataset}

We have come a long way since the days when a news article consisted
solely of a few paragraphs of text: online articles are illustrated by
pictures and even videos, and readers can share their comments on the
story in the same web-page, complementing the original document.

\begin{table*}[t!] 
\begin{center}
\begin{tabular}{|c|}\hline
\cellcolor[gray]{.9}{\bf News article tags} \\ \hline \hline
\multicolumn{1}{|m{17.7cm}|}{World news, UK news, Business, Politics, Society, Australia, Life and 
style, Sport, United States, World, Europe, Health, Entertainment, 
Australian politics, Nation, Football, Culture, Middle East and North 
Africa, Asia Pacific, Comment, Media, Sports, Law, Soccer, 
Environment, Africa, Labour, Conservatives, Russia, Blogposts, 
London, David Cameron, Features, Australia news, Technology, 
Money, Scotland, China, Food and drink, Ukraine, Education, NHS, 
Women, Sci/Tech, Tony Abbott, US politics, European Union, 
Editorial, Iraq, Economics, Barack Obama, Books, US news, Syria, 
Children, Music, Religion, Premier League, Film, Top Stories, Family,
Scottish independence, US foreign policy, Science, Ed Miliband, 
Crime, Liberal Democrats, France, England, Internet, Retail industry,  
Israel, Television, Race issues, Labor party, Police, Economic policy,  
New South Wales, Gender, Victoria, Vladimir Putin, Local 
government, UK Independence party (Ukip), Scottish politics,
Americas}\\
\hline
\end{tabular}
\caption{Some popular tags associated to news articles in the dataset.}
\label{tab:tags}
\end{center}
\end{table*}

\begin{table*}[tb]
\begin{center}
\begin{small}
\begin{tabular}{|l|cccccccc|}
\hline
\multicolumn{1}{|l}{\cellcolor[gray]{.9} \bf \scriptsize Source} & \multicolumn{1}{C{0.055\textwidth}}{\cellcolor[gray]{.9} \scriptsize \bf num. articles} & \multicolumn{1}{C{0.079\textwidth}}{\cellcolor[gray]{0.9} \scriptsize \bf avg. len. article} & \multicolumn{1}{C{0.09\textwidth}}{\cellcolor[gray]{.9} \scriptsize \bf avg. num. images} & \multicolumn{1}{C{0.07\textwidth}}{\cellcolor[gray]{.9} \scriptsize \bf avg. len. caption} & \multicolumn{1}{C{0.09\textwidth}}{\cellcolor[gray]{.9} \scriptsize \bf avg. num. comments} & \multicolumn{1}{C{0.07\textwidth}}{\cellcolor[gray]{.9} \scriptsize \bf avg. len. comment} & \multicolumn{1}{C{0.101\textwidth}}{\cellcolor[gray]{.9} \scriptsize \bf avg. num. shares} & \multicolumn{1}{C{0.06\textwidth}|}{\cellcolor[gray]{.9} \scriptsize \bf \% geo-located} \\ \hline\hline
{\bf \scriptsize Yahoo News}       & \footnotesize{10,834} & \footnotesize{$521 \pm 338$} & \footnotesize{$1.00 \pm 0.00$} & \footnotesize{$40 \pm 33$} & \footnotesize{$126 \pm 658$} & \footnotesize{$39 \pm 71$} & \footnotesize{n/a} & \footnotesize{$65.2\%$} \\ %news.yahoo.com
{\bf \scriptsize BBC News }         & \footnotesize{17,959} & \footnotesize{$380 \pm 240$} & \footnotesize{$1.54 \pm 0.82$} & \footnotesize{$14 \pm 4$}  & \footnotesize{$7 \pm 78$} & \footnotesize{$48 \pm 21$} & \footnotesize{n/a} &\footnotesize{ $48.7\%$} \\ %www.bbc.co.uk
{\bf \scriptsize The Irish Independent }    & \footnotesize{4,073}  & \footnotesize{$555 \pm 396$} & \footnotesize{$1.00 \pm 0.00$ }& \footnotesize{$14 \pm 14$} & \footnotesize{$1 \pm 6$} & \footnotesize{$17 \pm 5$} & \footnotesize{$4 \pm 20$} & \footnotesize{$52.3\%$} \\ %www.independent.ie
{\bf \scriptsize Sydney Morning Herald}         & \footnotesize{6,025}  & \footnotesize{$684 \pm 395$} & \footnotesize{$1.38 \pm 0.71$} & \footnotesize{$14 \pm 10$} & \footnotesize{$6 \pm 37$} & \footnotesize{$58 \pm 55$} & \footnotesize{$718 \pm 4976$} & \footnotesize{$60.4\%$} \\ %www.smh.com.au
{\bf \scriptsize The Telegraph}    & \footnotesize{29,757} & \footnotesize{$700 \pm 449$} & \footnotesize{$1.01 \pm 0.12$} & \footnotesize{$16 \pm 8$}  & \footnotesize{$59 \pm 251$} & \footnotesize{$45 \pm 65$} & \footnotesize{$355 \pm 2867$} & \footnotesize{$59.3\%$} \\ %www.telegraph.co.uk
{\bf \scriptsize The Guardian}    & \footnotesize{20,141} & \footnotesize{$786 \pm 527$} & \footnotesize{$1.18 \pm 0.59$} & \footnotesize{$20 \pm 8$}  & \footnotesize{$180 \pm 359$} & \footnotesize{$53 \pm 64$} & \footnotesize{$1509 \pm 7555$} & \footnotesize{$61.5\%$} \\ %www.theguardian.com
{\bf \scriptsize The Washington Post} & \footnotesize{9,839}  & \footnotesize{$777 \pm 477$} & \footnotesize{$1.10 \pm 0.43$} & \footnotesize{$25 \pm 17$} & \footnotesize{$98 \pm 342$} & \footnotesize{$43 \pm 50$} & \footnotesize{n/a} & \footnotesize{$61.3\%$} \\ %www.washingtonpost.com
\hline
\end{tabular}
\end{small}
\caption{BreakingNews dataset statistics. Mean and standard deviation, usually rounded to the nearest integer. See Section~\ref{sec:dataset} for description.}
\label{tab:sources}
\end{center}
\end{table*}

Studying the effects and interactions of these multiple modalities has
clear interesting applications, such as easing the work of journalist
by automatically suggesting pictures from a repository, or
determining the best way to promote a given article in order to reach
the widest readership. Yet, no benchmarks that capture this
multi-modality are available for scientific research. 

For these reasons, we propose a novel dataset of news articles with
images, captions, geolocation information and comments, which we will use to evaluate the previously described CNN and LSTM architectures on a variety of tasks. Our CNNs are based on the most recent state-or-the-art approaches in deep learning, and thus, this dataset is intended to be a touchstone to explore the current limits of these methodologies, which have been shown to be very effective when dealing with images associated with visually descriptive text.

\subsection{Description of the Dataset}
\label{subsec:datasetdescription}
The dataset consists of approximately 100,000 articles published
between the 1st of January and the 31th of December of 2014. All
articles include at least one image, and cover a wide variety of
topics, including sports, politics, arts, healthcare or local
news. Table~\ref{tab:tags} shows some of the most popular topics.

The main text of the articles was downloaded using the IJS
newsfeed~\cite{TrampuSIKDD2012}, which provides a
clean stream of semantically enriched news articles in multiple
languages from a pool of \textit{rss} feeds.

We restricted the articles to those that were written in English,
contained at least one image, and originated from a shortlist of
highly-ranked news media agencies (see Table~\ref{tab:sources}) to
ensure a degree of consistency and quality.  Given the geographic
distribution of the news agencies, most of the dataset is made of news stories in English-speaking countries in general, and the
United Kingdom in particular.  

For each article we downloaded the images, image captions and user
comments from the original article web-page. News article images are
quite different from those in existing captioned images
datasets like Flickr8K~\cite{HodoshJAIR2013} or MS-COCO~\cite{LinECCV2014}: often include close-up views of a person (46\% of the
pictures in BreakingNews contain faces) or complex
scenes. Furthermore, news image captions use a much richer vocabulary
than in existing datasets
(e.g.\ Flickr8K  has a total of 8,918 unique tokens,
while eight thousand random captions from BreakingNews already have
28,028), and they rarely describe the exact contents of the picture.

We complemented the original article images with additional pictures
downloaded from Google Images, using the full title of the article as
search query.  The five top ranked images of sufficient size in each
search were downloaded as potentially related images (in fact, the
original article image usually appears among them).

Regarding measures of article popularity, we downloaded all comments in the article page and the number of shares on different social networks (e.g.~Twitter, Facebook, LinkedIn) if this information was available.  Whenever possible, in addition to the full text of the comments, we  recovered the thread structure, as well as the author, publication date, likes (and dislikes) and number of replies. Since there were no share or comments information available for "The Irish Independent", we searched Twitter using the full title of the articles, and collected the tweets with links to the original article, or that mentioned a name associated with the newspaper (e.g.~@Independent\_ie, Irish Independent, @IndoBusiness) in place of comments. We considered the collective number of re-tweets as shares of the article. In this work, we have used the number of comments that is reported in the article web-page as the popularity measure, but many other options are possible  (e.g.~number of shares, number of comments by different users or number of comments of a certain length).

The IJS Newsfeed annotates the articles with geolocation information both for the news agency and for the article content. This information is primarily taken from the provided RSS summary, but sometimes it is not available and then it is inferred from the article using heuristics such as the location of the publisher, TLD country, or the story text. Fig. \ref{fig:geoloc_gt} shows a distribution of news story geolocation. 

Finally, the dataset is annotated for convenience with shallow and deep linguistic features (e.g. part of speech tags, inferred semantic topics, named entity detection and resolution, sentiment analysis) with the \textit{XLike}~\footnote{\url{http://www.xlike.org/language-processing-pipeline/}} and the \textit{Enrycher}~\footnote{\url{http://ailab.ijs.si/tools/enrycher/}} NLP pipelines. 

\subsection{Comparison to similar datasets}
To the best of our knowledge, the only publicly available multi-modal
news datasets similar to BreakingNews  is the BBC News~\cite{FengPAMI2013}, which contains a total of 3,361
articles with images and captions.  This size is clearly insufficient
to train data-hungry models like LSTM recurrent neural networks,
that are the state-of-the-art techniques for caption generation.
Furthermore, in addition to the difference in size, the BBC News
does not include geolocation information, tags, social
network shares or user comments, and all articles come from a single
source.
% % % % % % % % % % % % % % % % % % % % % % % % % % % % % % % % % %

\section{Results}
\label{sec:results}

In this section, we present experimental results on the five tasks we considered. We first discuss CNN results for source detection, geolocation and popularity prediction and article illustration, followed by a discussion on LSTM and mixed LSTM/CNNs model for caption generation.

\subsection{Implementation Details}
\label{sec:implementation}

Before describing the experimental results, we discuss the  technical aspects related to the preparation of dataset and  to the implementation details, including the hyper-parameters set-up for representation and learning.

\vspace{2mm}
{\noindent \bf Dataset considerations for the experimental setup:} The BreakingNews dataset is split into train, validation and test sets with 60\%, 20\% and 20\% articles respectively.  In order to ensure fairness in the experiments, we also checked there was almost no overlap of images between the sets using the VGG19 features and a cosine distance. This ratio of near-identical image pairs  was in the order of $10^{-6}$. The measure of article popularity that we have used corresponds to the number of comments, as reported in the article web-page.

\vspace{2mm}
{\noindent \bf Textual representations:} In total there are 44,665 unique tokens in the training set, so the BoW representation is $D_b=$ 44,665 dimensional. For Word2Vec, we used the BreakingNews training set with the skip-gram method to learn the embedding space. The size of Word2vec embedding space was $D_w=500$, the window size was 30, the sampling rate was set to 1e-5, and the number of negative samples was set to 5. These hyper-parameters were chosen based on the article illustration task for a small subset. We also experimented with a publicly available Word2Vec model trained with the Google 100 billion words dataset, but the performance was worse. 

\vspace{2mm}
{\noindent \bf CNNs for article analysis:}
The proposed CNN architecture including our novel GCD  and CCA losses were implemented using the Caffe framework~\cite{jia14arxiv_caffe}. Unless specified otherwise, in the experiments we used a base learning rate of 0.05, which dropped by a factor of 0.1 after every 1,000 iterations. Stochastic gradient descent was employed as the solver, for which the momentum was set to 0.9. The regularisation parameter weight decay was set to 0.0005. For most tasks a larger batch size tended to lead to better performance. The results reported in this paper were obtained with a batch size of $m=64$. All our experiments were performed on an NVIDIA Tesla K40C GPU with 12G memory. It takes approximately 10 hours for training to converge. Depending on task, the total number of parameters learnt in the CNN ranges from approximately 0.7M to 1.2M.

\vspace{2mm}
{\noindent \bf LSTMs for caption generation: } The two architectures in Fig.~\ref{fig:cnn_arch}c,d were implemented using Neuraltalk2\footnote{https://github.com/karpathy/neuraltalk2}. We set both the input encoding size and LSTM memory size to 256, the learning rate to 0.0002, the batch size to 4, and kept all other parameters default. On the NVIDIA K40C GPU it typically took several days for the training to converge. The total number of parameters  was approximately 50M  and 200M for the two architectures respectively.

\begin{figure*}[t!]
\centering
\subfloat[BoW TF-IDF]   {\includegraphics[trim= 5 0 5 0,clip,height=42mm,width=59mm]{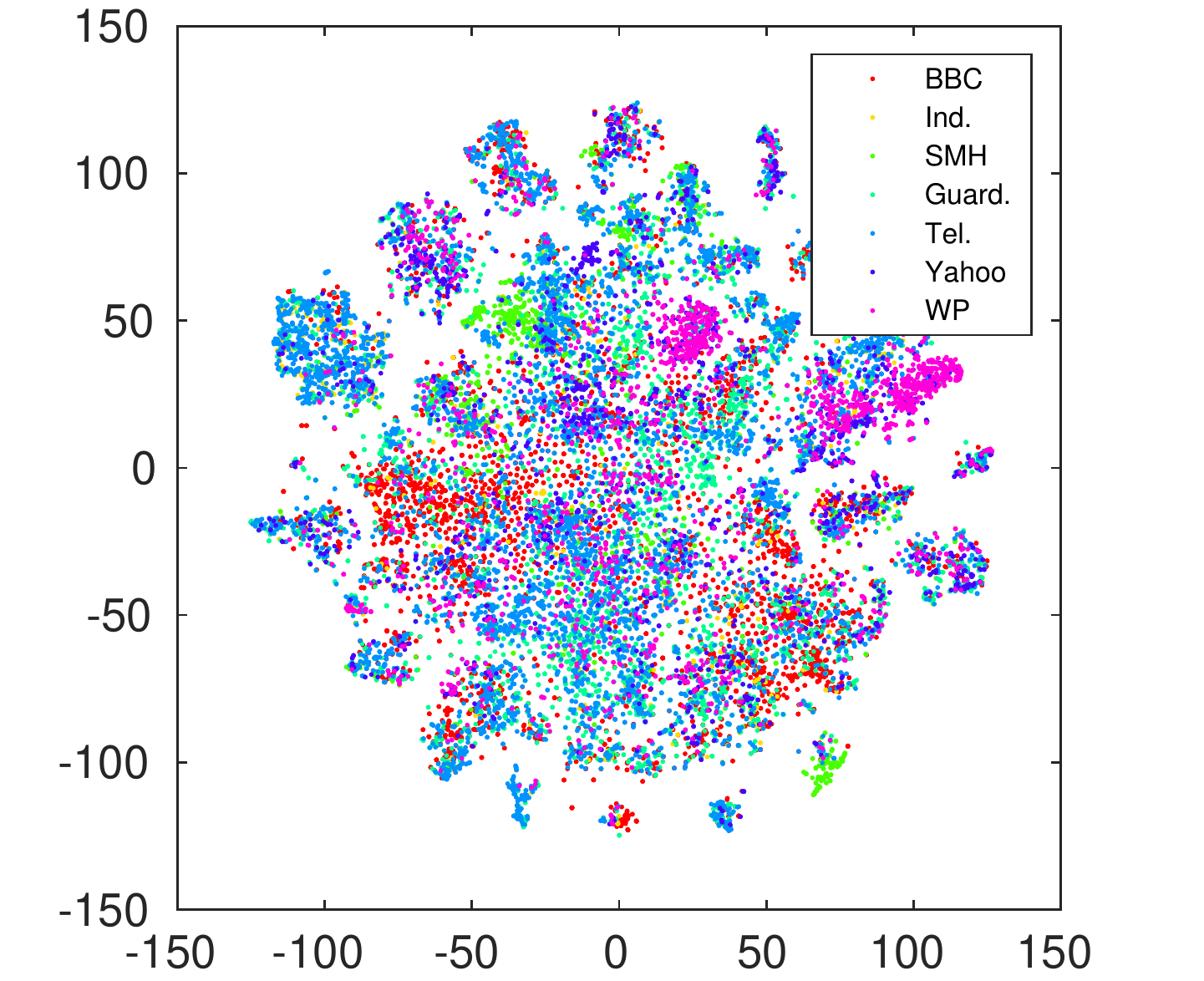}} \hspace{-1mm}
\subfloat[W2V mean]     {\includegraphics[trim= 5 0 5 0,clip,height=42mm,width=59mm]{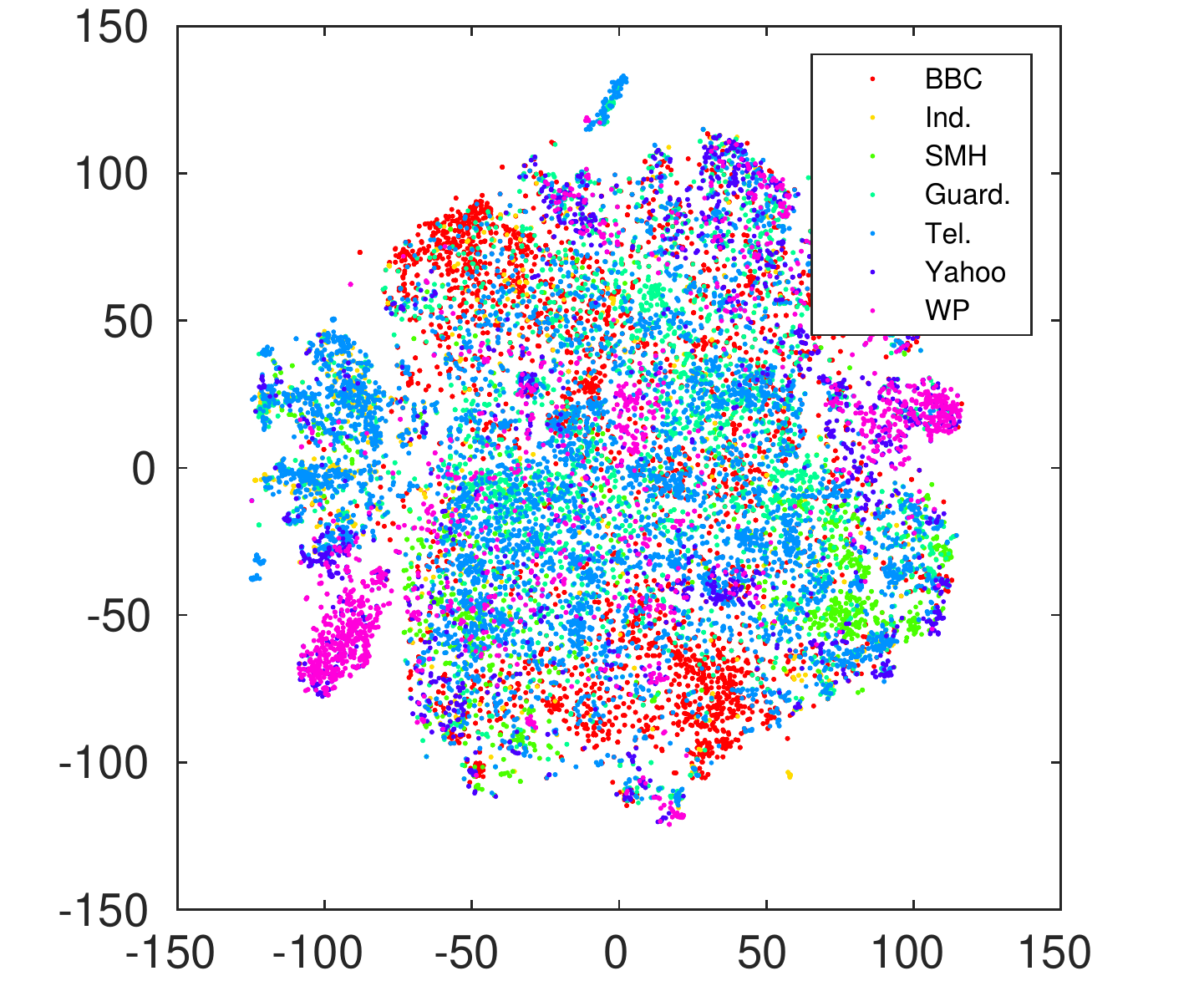}} \hspace{-1mm}
\subfloat[CNN FCo layer]{\includegraphics[trim= 5 0 5 0,clip,height=42mm,width=59mm]{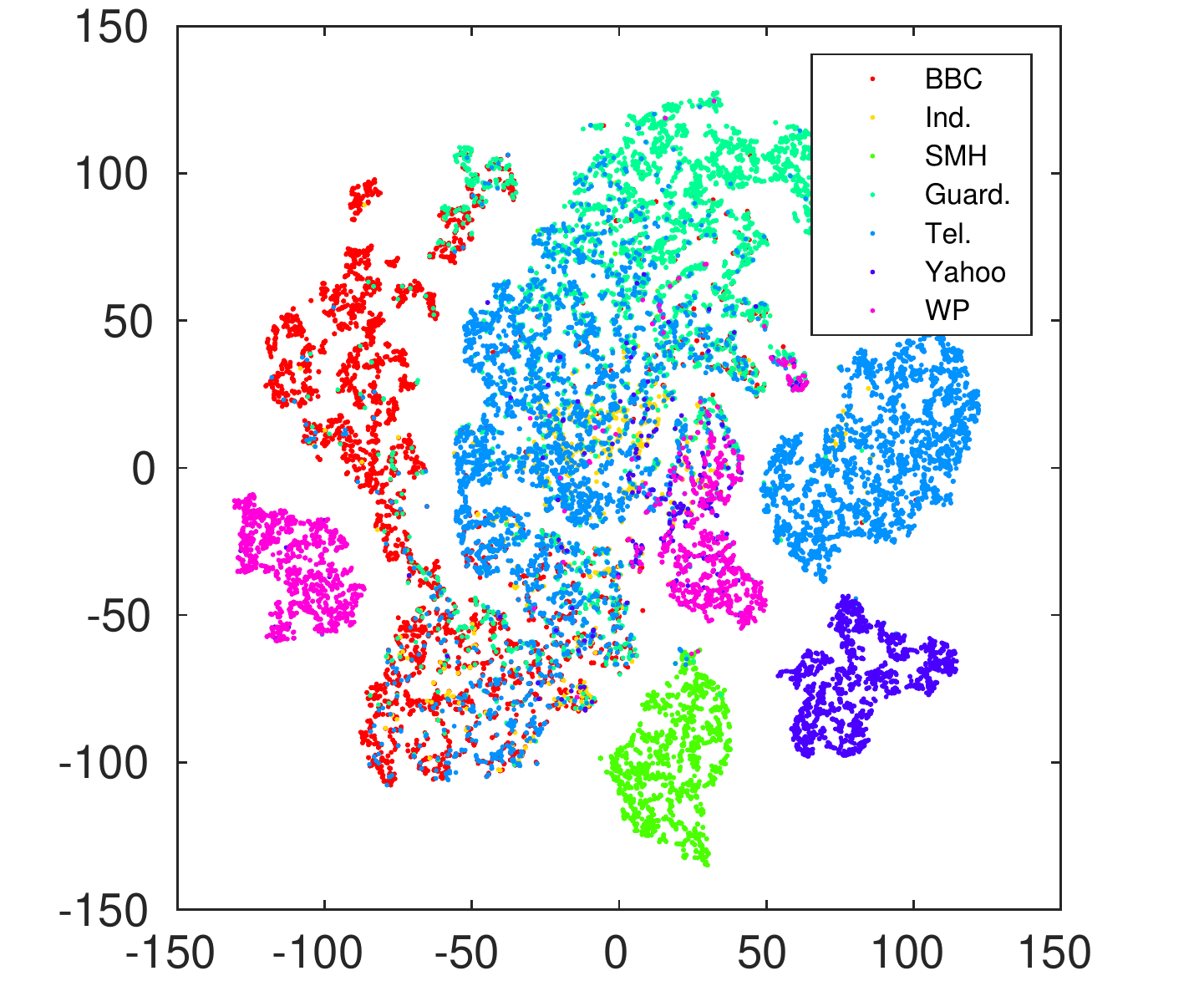}}
\vspace{0mm}
\caption{Source detection:  t-SNE embedding of shallow and deep features for the 19,849 articles from the test set.}
\label{fig:source_tsne}
\end{figure*}

\subsection{CNNs for article analysis}
\label{subsec:results:cnn}

{\noindent \bf Source detection: } For this task we compare the proposed CNN trained with the full $500\times 8,000$ Word2Vec matrix with the popular shallow learner linear SVM. For the latter, we consider four versions of shallow textual features: the full 44,665 dimensional TF-IDF weighted BoW feature; its truncated version with 5,000 dimensions; the mean of Word2Vec matrix; and the max of Word2Vec matrix (direct usage of the full matrix was discarded right away as impractical and not robust to sentence variations). 
Additionally, note that the source detection problem is unbalanced, as can be seen in Table \ref{tab:sources}. We therefore use two performance metrics: the overall accuracy, and the balanced accuracy which is the average of per-class accuracies.

Results in Table~\ref{table:result:source} show that the full BoW feature outperforms its truncated version, and the mean of Word2Vec is much better than the max version. When comparing the two types of features, BoW's small advantage may be attributed to its much higher dimensionality (44,665/5,000 vs. 500). Finally, the proposed CNN architecture produces accuracies more than 10 points higher than the best shallow feature. In Fig.~\ref{fig:source_tsne} we plot the 2D t-SNE~\cite{maaten08tsne} embedding of the shallow features and the output of the {\em FCo} layer of the CNN for the test set, where the points are colour-coded with class labels. It is evident that the deeply-learnt representation in CNN provides a much better separation of the different classes.

\begin{table}
\renewcommand{\arraystretch}{1.3}
\centering
\begin{small}
\begin{tabular}{|cccc|}
\hline
\multicolumn{4}{|c|}{\cellcolor[gray]{.9} \bf Source Detection} \\ \hline\hline
  &  &  {\bf Accuracy}  &  {\bf Bal. accuracy}   \\
%\noalign{\vskip 1.0mm}
\hline
%\noalign{\vskip 1.0mm}
\multirow{4}{*}{SVM}        & BoW TF-IDF         &      72.9  &      72.6  \\
                            & BoW TF-IDF 5000-d  &      70.8  &      70.0  \\
                            & W2V mean           &      68.6  &      64.4  \\
                            & W2V max            &      57.3  &      53.4  \\
% \noalign{\vskip 0.5mm}
\hline
% \noalign{\vskip 0.5mm}
                       CNN  & W2V matrix       & {\bf 81.2} & {\bf 80.7} \\ \hline
\end{tabular}
\end{small}
\caption{Results of source detection. Due to the unbalance between the classes, balanced accuracy is also reported. CNN shows a clear advantage over shallow learners for this task.}
\label{table:result:source}
\end{table}

\begin{table}[t!]
\renewcommand{\arraystretch}{1.3}
\centering
\begin{small}
\begin{tabular}{|cccccc|}
\hline
\multicolumn{6}{|c|}{\cellcolor[gray]{.9}{\bf Article Illustration}} \\
\hline\hline
\multicolumn{6}{|c|}{\cellcolor[gray]{.96}{\bf Text to image}} \\
\hline
{} &{\bf Text} & {\bf Image} & {\bf R@1}  & {\bf R@10}   &  {\bf MR}   \\
\multirow{6}{*}{CCA} & \multirow{3}{*}{BoW TF-IDF} & {\footnotesize VGG19}        &       5.7 &      14.6  &      448  \\
                     &                             & {\footnotesize Places}       &       5.1 &      12.1  &      801  \\
                     &                             & {\footnotesize VGG19+Places} & {\bf 6.0} & {\bf 15.3} & {\bf 397} \\
%\noalign{\vskip 1.0mm}
                     & \multirow{3}{*}{W2V mean}   & {\footnotesize VGG19}        &       2.4 &       9.6  &      499  \\
                     &                             & {\footnotesize Places}       &       2.2 &       7.9  &      820  \\
                     &                             & {\footnotesize VGG19+Places} &       3.5 &      11.9  &      415  \\
%\noalign{\vskip 0.5mm}
\hline
%{} &{\bf Text} & {\bf Image} & {\bf R@1}  & {\bf R@10}   &  {\bf MR}   \\
%\noalign{\vskip 0.5mm}
CNN                  & W2V matrix                  & {\footnotesize VGG19+Places} &       3.3  &      12.2 &      405  \\
%\noalign{\vskip 1.0mm}
\hline\hline
%\noalign{\vskip 1.0mm}
\multicolumn{6}{|c|}{\cellcolor[gray]{.96}{\bf Text to related images}} \\
%\noalign{\vskip 1.0mm}
\hline
{} &{\bf Text} & {\bf Image} & {\bf R@1}  & {\bf R@10}   &  {\bf MR}   \\
%\noalign{\vskip 1.0mm}
\multirow{3}{*}{CCA} & \multirow{3}{*}{W2V mean}   & {\footnotesize VGG19}        &       2.9 &      14.5  &      161  \\
                     &                             & {\footnotesize Places}       &       2.4 &      11.8  &      257  \\
                     &                             & {\footnotesize VGG19+Places} & {\bf 3.7} & {\bf 16.7} &      {\bf 137}  \\
%\noalign{\vskip 1.0mm}
\hline\hline
%\noalign{\vskip 1.0mm}
\multicolumn{6}{|c|}{\cellcolor[gray]{.96}{\bf Text to image+caption}} \\
%\noalign{\vskip 1.0mm}
\hline
{} &{\bf Text} & {\bf Image} & {\bf R@1}  & {\bf R@10}   &  {\bf MR}   \\
%\noalign{\vskip 1.0mm}
\multirow{2}{*}{CCA} & \multirow{2}{*}{W2V mean} & {\footnotesize VGG19+Orig.} & 25.0  & 55.6 &    7 \\
                     &                           & {\footnotesize VGG19+Anon.} & 11.8  & 33.2 & 42 \\ \hline
\end{tabular}
\end{small}
\caption{Article Illustration experiments. Three  setups to evaluate  different alternatives (learning approaches and features) for automatically illustrating images, and exploring the problem.}
\label{table:result:illustration}
\end{table}

\vspace{2mm}
{\noindent \bf Article illustration: } We compare the CNN with the Canonical Correlation Analysis (CCA), which is a standard shallow learner for text and image matching~\cite{GongECCV2014,HodoshJAIR2013}. Given textual and visual representations for a pair of article and image, CCA finds the optimal projection that maximises the correlation between the two modalities. For textual representation on the CCA we again consider both BoW and Word2Vec mean; and for image representation we use activations of the pretrained VGG19 and Places models, and the concatenation of the two. Recall that for  the CNN we use  the full  $500\times 8,000$ Word2Vec representation for the text,  and the 8,192-dimensional representation for the images, resulting from the concatenation of the VGG19 and Places activations.

For each test article we rank the 23,200 test images according to the projection learnt in CCA or CNN, and measure the performance in terms of recall of the ground truth image at the $k^{\textrm {th}}$ rank position (R@k), and the median rank (MR) of the ground truth. Note that a higher R@k or a lower MR score indicates a better performance. 

Results in the top  of Table~\ref{table:result:illustration} demonstrate that when CCA is used, with either textual representation the VGG19-based visual feature outperforms Places-based feature by a significant margin; while combining the two further improves the performance. Comparing the two textual representations, BoW with TF-IDF again has an edge over the mean of Word2Vec.

The CNN, although using the full Word2Vec matrix and VGG19+Places feature as input, only manages to marginally outperform its shallow counterpart (CCA with Word2Vec mean and VGG19+Places). We hypothesize that this is because in news articles, the text is only loosely related to the content of the associated image. This is in contrast to standard image/text matching datasets such as MS-COCO~\cite{LinECCV2014} and Flickr30k~\cite{GongECCV2014}, where the associated caption explicitly describes the content of the image. In any event, note that the MR values are around $400$, which, considering a pool of more than $24$K images, are very remarkable results.

In order to better understand the nature of the problem  and the   challenges of the proposed dataset we  performed  two additional experiments. For efficiency's sake, in these experiments we just considered the CCA, which can be trained much faster (requires computing an SVD of a matrix of size the number of training images) than the back-propagation scheme needed for the CNN.

In the first setting, the related images collected from the Internet replace the original image associated with the article. Visual features are extracted from the related images and averaged as the visual representation. The results in  Table~\ref{table:result:illustration}-middle indicate a large performance improvement. For example, the MR decreases from 499 to 161 for the W2V mean/VGG19 feature combination. This shows that the original images are weakly correlated with the actual articles, and this correlation is improved by considering additional Internet images.

Along the same lines, in the second setting we assumed the images were already equipped with captions. The W2V embeddings were computed for each caption and the mean was concatenated with the visual feature as the final "visual" representation. We considered two variants: one with the original captions and an "anonymized" one, with proper nouns replaced by common nouns, i.e., person names  replaced by \emph{someone}, place names by \emph{somewhere} and organizations by \emph{organization}. The results in the bottom  of Table~\ref{table:result:illustration} show that both variants substantially improve the performance,  reducing MR from 499 to 7 and 42, respectively. Again, this confirms that the main challenge of the BreakingNews dataset is the lack of correlation between  images and  articles. While the results obtained considering the original images are very promising, these last two experiments indicate there is still a big margin for improvement.

\begin{table} [t!]
\renewcommand{\arraystretch}{1.3}
\centering
\begin{small}
\begin{tabular}{|cccc|}
\hline
\multicolumn{4}{|c|}{\cellcolor[gray]{.9} \bf Popularity Prediction} \\
\hline\hline
 & &  {\bf Mean $\ell_1$} &  {\bf Median $\ell_1$}   \\
% \noalign{\vskip 1.0mm}
\hline
%\noalign{\vskip 1.0mm}
\multirow{2}{*}{SVR}  & BoW TF-IDF       &     70.25 &      6.73 \\
                      & W2V mean         & \bf 69.75 &      3.63 \\
%\noalign{\vskip 0.5mm}
\hline
%\noalign{\vskip 0.5mm}
CNN                   & W2V matrix       &     70.72 &  \bf 1.09 \\
\hline
\end{tabular}
\end{small}
\caption{Results of the popularity prediction task. The mean and median distance between the predicted and actual number of comments is shown. Given the long tail nature of the distribution of user comments, the median is more significant.}
\label{table:result:popularity}
\end{table}

\vspace{2mm}
{\noindent \bf Popularity prediction: } For this task we used Support Vector Regression (SVR) as the baseline method. Performance was measured as mean and median of $\ell_1$ distance between the predicted popularity (number of comments) and the ground truth. Results in Table~\ref{table:result:popularity} show that with median $\ell_1$ distance, CNN is significantly better than SVR with Word2Vec mean, which in turn outperforms SVR with BoW features. When mean $\ell_1$ distance is used as metric the performance difference between the methods is less pronounced. However, given the large variation in the number of comments (e.g. many articles have zero comments while some have as many as 14,298), the median is a more stable performance metric.

\vspace{2mm}
{\noindent \bf Geolocation prediction: } We again used SVR as a baseline for this 2D regression problem. Another baseline was the CNN but with an Euclidean loss, as opposed to the GCD loss. In the datasets there are articles with multiple geolocations. For training, we only used the first geolocation, while for testing we computed the GCD as defined in Eq.~\eqref{gcd} between the predicted geolocation and the nearest ground truth.

Table~\ref{table:result:geolocation} reports mean and median GCD of various learning schemes. CNN with the proposed GCD loss has a clear advantage over not only the shallow learners, but also CNN with Euclidean loss. This again confirms the effectiveness of the proposed CNN architecture, especially when coupled with a proper loss function. It is also interesting to note that although using different inputs (text vs. image) and different approaches, the median GCD achieved in our work ($900$ kilometers of error) is comparable to that of the recent work~\cite{WeyandARXIV2016}.

\begin{table}
\renewcommand{\arraystretch}{1.3}
\centering
\begin{small}
\begin{tabular}{|cccc|}
\hline
\multicolumn{4}{|c|}{\cellcolor[gray]{.9} \bf Geolocation Prediction} \\
\hline\hline
  & &  {\bf Mean GCD} &  {\bf Median GCD}   \\
% \noalign{\vskip 1.0mm}
\hline
% \noalign{\vskip 1.0mm}
\multirow{2}{*}{SVR}  & BoW TF-IDF       &     2.87 &     1.72 \\
                      & W2V mean         &     2.52 &     1.37 \\
% \noalign{\vskip 0.5mm}
\hline
% \noalign{\vskip 0.5mm}
CNN Euc               & W2V matrix       &     2.40 &     1.42 \\
CNN GCD               & W2V matrix       & \bf 1.92 & \bf 0.90 \\
\hline
\end{tabular}
\end{small}
\caption{Results of the geolocation prediction task. Results are expressed in thousands of kilometres. Deep learning performs better at this task, and the Great Circle Distance (GCD) is a better objective than the Euclidean distance.}
\label{table:result:geolocation}
\end{table}

\begin{table}
\renewcommand{\arraystretch}{1.3}
\centering
\begin{small}
\begin{tabular}{|ccccc|}
\hline
\multicolumn{5}{|c|}{\cellcolor[gray]{.9} \bf Single task vs Multitask vs Transfer Learning} \\
\hline\hline
&                                     &  {\bf Geolocation} &  {\bf Popularity}    &  {\bf Source} \\
&                                     &  {\bf Median GCD}  &  {\bf Median $\ell_1$}  &  {\bf Bal. acc.}  \\
%\noalign{\vskip 1.0mm}
\hline
%\noalign{\vskip 1.0mm}
\multicolumn{2}{|c}{Single-task}       &  {\bf 0.90}  &  1.09             & {\bf 80.7}  \\
%\noalign{\vskip 0.5mm}
\hline
%\noalign{\vskip 0.5mm}
\multirow{3}{*}{Multitask}    & G+P   &  1.17        &  1.68             &  -          \\
                              & G+S   &  1.22        &  -                &  79.1       \\
                              & P+S   &  -           &  1.94             &  79.1       \\
%\noalign{\vskip 0.5mm}
\hline
%\noalign{\vskip 0.5mm}
\multirow{6}{*}{Transfer}     & G$\rightarrow$P  &  -           &  1.16             &  -          \\
                              & G$\rightarrow$S  &  -           &  -                &  80.2       \\
                              & P$\rightarrow$G  &  0.97        &  -                &  -          \\
                              & P$\rightarrow$S  &  -           &  -                &  77.6       \\
                              & S$\rightarrow$G  &  0.92        &  -                &  -          \\
                              & S$\rightarrow$P  &  -           &  {\bf 0.63}       &  -          \\
\hline
\end{tabular}
\end{small}
\caption{Comparing single-task, multitask and transfer learing. G: geolocation; P: popularity; S: source. An arrow shows the direction of transfer, for example, G$\rightarrow$P means trained on task G and transferred to task P.}
\label{table:result:st_vs_mt_vs_tt}
\end{table}

\vspace{2mm}
{\noindent \bf Single-task, multitask and transfer learning: } In this section, we compare single-task, multitask and transfer learning on the three scenarios where CNN achieved the best results, namely, source detection and geolocation and popularity  prediction. For multitask, we jointly train a model with each of the three pairs of tasks. For transfer learning, we train six models using the three pairs and in both directions.

The results in Table~\ref{table:result:st_vs_mt_vs_tt} show that single-task learning tends to have the best performance, and that transfer learning in general outperforms multitask learning. The fact that multitask learning does not perform as well as the other alternatives  seems to indicate that our tasks are not as "compatible" as those considered in~\cite{zhang14eccv}. Forcing the lower layers of the CNN to share common representations harms the performance of both tasks. We experimented with CNN architectures that further decouple the tasks but did not observe any improvements.

On the other hand, in most cases the performance of transfer learning is comparable to that of single-task learning. When transferring from "source" to "popularity" prediction, the performance is even higher than single-task "popularity" prediction. One possible explanation is that training on one task helps to learn low and mid-level representations that better generalize for another task. Note that in both multitask and transfer learning, we set the learning rate of the \emph{FCo} layers to 1/10 of the base learning rate. Allowing the \emph{FCo} layers to change slowly helps reducing the co-adaptation effect~\cite{YosinskiNIPS2014} and tends to find a better solution.

\subsection{LSTM for Caption generation}
\label{subsec:results:lstm}

In Table~\ref{table:result:caption} we report caption generation results with the two models in Fig.~\ref{fig:cnn_arch}, i.e., LSTM with fixed feature and end-to-end CNNs/LSTM, respectively. The results indicate that combining textual and visual features, and moreover combining more types of visual features (VGG19 and Places) helps to improve caption generation. On the other hand, finetuning the CNN features in the end-to-end model does not improve the results. We hypothesize that this is again due to the low correlation between the caption and the content of the image. It should also be noted that the performance of all models is poor compared to that achieved by very similar CNN models in standard image caption generation tasks~\cite{GongECCV2014,LinECCV2014}. Again, we argue that captions of news images are of very different nature. Generating such captions is considerably more challenging and requires more careful and specific treatment and evaluation metric, than that done by current state-of-the-art approaches. The proposed BreakingNews dataset, therefore, poses new challenges for future research.

\begin{table*}
\renewcommand{\arraystretch}{1.3}
\centering
\begin{small}
\begin{tabular}{|cccccccc|}
\hline
\multicolumn{8}{|c|}{\cellcolor[gray]{.9} \bf Caption Generation Results} \\
\hline\hline
                            & {\bf Text}        & {\bf Image}          &     {\bf METEOR}  &  {\bf BLEU-1}    &  {\bf BLEU-2}   &   {\bf BLEU-3}  &  {\bf BLEU-4}  \\
%\noalign{\vskip 1.0mm}
\hline
%\noalign{\vskip 1.0mm}
\multirow{5}{*}{LSTM w/ fixed feat.}  & W2V mean    & -              &       5.2  &      16.3  &      6.6  &      3.2  &     1.7  \\
                            & -           & VGG19          &       3.9  &      14.4  &      4.8  &      1.8  &     0.8  \\
                            & -           & VGG19+Places   &       4.3  &      14.4  &      4.9  &      2.0  &     1.0  \\
                            & W2V mean    & VGG19          &       4.9  &      16.2  &      6.5  &      3.1  &     1.6  \\
                            & W2V mean    & VGG19+Places   &   \bf{5.3} &      17.2  &      7.0  &  \bf{3.4} & \bf{1.9} \\
%\noalign{\vskip 1.0mm}
\hline
%\noalign{\vskip 1.0mm}
CNNs + LSTM                 & W2V         & VGG19          &       5.2  &  \bf{19.6} &  \bf{8.9} &      1.6  &     0.5  \\
\hline
\end{tabular}
\end{small}
\caption{Results of caption generation. Performance is overall low, which highlights the difficulty of the task, due to the subtle and high-level relation between the images and the captions.}
\label{table:result:caption}
\end{table*}

\section{Conclusion}

In this paper we have introduced new deep CNN architectures to combine weakly correlated text and image representations and address several  tasks in the domain of news articles, including story illustration, source detection, geolocation and popularity prediction, and automatic captioning. In particular, we propose an adaptive CNN architecture that shares most of its structure for all the tasks. Addressing each problem then requires designing specific loss functions, and we introduce a metric based on the Great Circle Distance for geolocation and Deep Canonical Correlation Analysis for article illustration. All these technical contributions are exhaustively evaluated on a new dataset, BreakingNews,  made of approximately 100K  news  articles (about 2 orders of magnitude more than similar existing datasets), and additionally including a diversity of metadata (like GPS coordinates and popularity metrics) that makes it possible to explore new problems. Overall results are very promising, specially for the tasks of source detection, article illustration and geolocation. The automatic caption generation task, however, is clearly more sensitive to loosely related text and images. Designing new metrics able to handle this situation, is part of our future work.

\section*{Acknowledgments}
This work has been partially funded by Spanish Ministry of Economy and Competitiveness under project RobInstruct  TIN2014-58178-R, by the ERA-net CHISTERA projects VISEN PCIN-2013-047 and I-DRESS PCIN-2015-147. The authors would like to thank Blaz Novak for his help with the IJS Newsfeed.

\bibliographystyle{plain}
\bibliography{references}

\begin{thebibliography}{10}

\bibitem{andrew13icml}
G.~Andrew, R.~Arora, J.~Bilmes, and K.~Livescu.
\newblock Deep canonical correlation analysis.
\newblock In {\em ICML}, 2013.

\bibitem{BandariARXIV2012}
R.~Bandari, S.~Asur, and B.~Huberman.
\newblock The pulse of news in social media: Forecasting popularity.
\newblock {\em arXiv preprint arXiv:1202.0332}, 2012.

\bibitem{BaoICWWW2013}
P.~Bao, H.~Shen, J.~Huang, and X.~Cheng.
\newblock Popularity prediction in microblogging network: a case study on sina
  weibo.
\newblock In {\em International conference on World Wide Web Companion}, pages
  177--178, 2013.

\bibitem{BarnardJMLR2003}
K.~Barnard, P.~Duygulu, D.~Forsyth, N.~De Freitas, D.~Blei, and M.~Jordan.
\newblock Matching words and pictures.
\newblock {\em The Journal of Machine Learning Research}, 3:1107--1135, 2003.

\bibitem{BarnardICCV2001}
K.~Barnard and D.~Forsyth.
\newblock Learning the semantics of words and pictures.
\newblock In {\em ICCV}, volume~2, pages 408--415. IEEE, 2001.

\bibitem{CaoICM2009}
L.~Cao, J.~Yu, J.~Luo, and T.~Huang.
\newblock Enhancing semantic and geographic annotation of web images via
  logistic canonical correlation regression.
\newblock In {\em ACM International Conference on Multimedia}, pages 125--134.
  ACM, 2009.

\bibitem{ChangSPONSOR2014}
A.~Chang, M.~Savva, and C.~Manning.
\newblock Interactive learning of spatial knowledge for text to 3d scene
  generation.
\newblock {\em Sponsor: Idibon}, page~14, 2014.

\bibitem{ChenCVPR2011}
D.~Chen, G.~Baatz, K.~K{\"o}ser, S.~Tsai, R.~Vedantham, T.~Pylv{\"a},
  K.~Roimela, X.~Chen, J.~Bach, M.~Pollefeys, et~al.
\newblock City-scale landmark identification on mobile devices.
\newblock In {\em CVPR}, pages 737--744. IEEE, 2011.

\bibitem{ChenCVPR2015}
X.~Chen and C.~Zitnick.
\newblock Mind's eye: A recurrent visual representation for image caption
  generation.
\newblock In {\em CVPR}, 2015.

\bibitem{CoelhoLNCS2012}
F.~Coelho and C.~Ribeiro.
\newblock {Image abstraction in crossmedia retrieval for text illustration}.
\newblock {\em Lecture Notes in Computer Science}, 7224 LNCS:329--339, 2012.

\bibitem{collobert11jmlr}
R.~Collobert, J.~Weston, L.~Bottou, M.~Karlen, K.~Kavukcuoglu, and P.~Kuksa.
\newblock Natural language processing (almost) from scratch.
\newblock {\em JMLR}, 12(08):2493--2537, 2011.

\bibitem{CoyneCCGIT2001}
B.~Coyne and R.~Sproat.
\newblock Wordseye: an automatic text-to-scene conversion system.
\newblock In {\em Conference on Computer Graphics and Interactive Techniques},
  pages 487--496. ACM, 2001.

\bibitem{CrandallICWWW2009}
D.~Crandall, L.~Backstrom, D.~Huttenlocher, and J.~Kleinberg.
\newblock Mapping the world's photos.
\newblock In {\em International Conference on World Wide Web}, pages 761--770.
  ACM, 2009.

\bibitem{DengCVPR2009}
J.~Deng, W.~Dong, R.~Socher, K.~Li, K.~Li, and L.~Fei-Fei.
\newblock Imagenet: A large-scale hierarchical image database.
\newblock In {\em CVPR}, pages 248--255, 2009.

\bibitem{DharCVPR2011}
S.~Dhar, V.~Ordonez, and T.~Berg.
\newblock High level describable attributes for predicting aesthetics and
  interestingness.
\newblock In {\em CVPR}, pages 1657--1664. IEEE, 2011.

\bibitem{DingICVLDB2000}
J.~Ding, L.~Gravano, and N.~Shivakumar.
\newblock Computing geographical scopes of web resources.
\newblock In {\em International Conference on Very Large Data Bases}, VLDB '00,
  pages 545--556, 2000.

\bibitem{DonahueCVPR2015}
J.~Donahue, L.~Hendricks, S.~Guadarrama, M.~Rohrbach, S.~Venugopalan, .~Saenko,
  and T.Darrell.
\newblock Long-term recurrent convolutional networks for visual recognition and
  description.
\newblock In {\em CVPR}, 2015.

\bibitem{DouzeCVPR2011}
M.~Douze, A.~Ramisa, and C.~Schmid.
\newblock Combining attributes and fisher vectors for efficient image
  retrieval.
\newblock In {\em CVPR}, pages 745--752, 2011.

\bibitem{EveringhamIJCV2010}
M.~Everingham, L.~Van Gool, C.~Williams, J.~Winn, and A.~Zisserman.
\newblock The pascal visual object classes (voc) challenge.
\newblock {\em International journal of computer vision}, 88(2):303--338, 2010.

\bibitem{FangCVPR2015}
H.~Fang, S.~Gupta, F.~Iandola, R.~Srivastava, L.~Deng, and P.~Doll{\'a}r
  others.
\newblock From captions to visual concepts and back.
\newblock In {\em CVPR}, pages 1473--1482, 2015.

\bibitem{FarhadiECCV2010}
A.~Farhadi, M.~Hejrati, M.~Sadeghi, P.~Young, C.~Rashtchian, J.~Hockenmaier,
  and D.~Forsyth.
\newblock Every picture tells a story: Generating sentences from images.
\newblock In {\em ECCV}, pages 15--29. Springer, 2010.

\bibitem{FengACL2010}
Y.~Feng and M.~Lapata.
\newblock {Topic Models for Image Annotation and Text Illustration}.
\newblock {\em Conference of the North American Chapter of the ACL: Human
  Language Technologies}, (June):831--839, 2010.

\bibitem{FengPAMI2013}
Y.~Feng and M.~Lapata.
\newblock Automatic caption generation for news images.
\newblock {\em PAMI}, 35(4):797--812, 2013.

\bibitem{GobronVC2010}
S.~Gobron, J.~Ahn, G.~Paltoglou, M.~Thelwall, and D.~Thalmann.
\newblock From sentence to emotion: a real-time three-dimensional graphics
  metaphor of emotions extracted from text.
\newblock {\em The Visual Computer}, 26(6-8):505--519, 2010.

\bibitem{GongECCV2014}
Y.~Gong, L.~Wang, M.~Hodosh, J.~Hockenmaier, and S.~Lazebnik.
\newblock Improving image-sentence embeddings using large weakly annotated
  photo collections.
\newblock In {\em ECCV}, pages 529--545. 2014.

\bibitem{GygliICCV2013}
M.~Gygli, H.~Grabner, H.~Riemenschneider, F.~Nater, and L.~Van Gool.
\newblock The interestingness of images.
\newblock In {\em ICCV}, pages 1633--1640, 2013.

\bibitem{HaysCVPR2008}
J.~Hays and A.~Efros.
\newblock Im2gps: estimating geographic information from a single image.
\newblock In {\em CVPR}, pages 1--8, 2008.

\bibitem{hochreiter97nc}
S.~Hochreiter and J.~Schmidhuber.
\newblock Long short-term memory.
\newblock {\em Neural Computation}, 9(8):1735--1780, 1997.

\bibitem{HodoshJAIR2013}
M.~Hodosh, P.~Young, and J.~Hockenmaier.
\newblock Framing image description as a ranking task: Data, models and
  evaluation metrics.
\newblock {\em Journal of Artificial Intelligence Research}, 47:853--899, 2013.

\bibitem{HuangCTAA2013}
C.~Huang, C.~Li, and M.~Shan.
\newblock {VizStory: Visualization of Digital Narrative for Fairy Tales}.
\newblock {\em Conference on Technologies and Applications of Artificial
  Intelligence}, pages 67--72, 2013.

\bibitem{IsolaPAMI2014}
P.~Isola, J.~Xiao, D.~Parikh, A.~Torralba, and A.~Oliva.
\newblock What makes a photograph memorable?
\newblock {\em PAMI}, 36(7):1469--1482, 2014.

\bibitem{jia14arxiv_caffe}
Y.~Jia, E.~Shelhamer, J.~Donahue, S.~Karayev, J.~Long, R.~Girshick,
  S.~Guadarrama, and T.~Darrell.
\newblock Caffe: Convolutional architecture for fast feature embedding.
\newblock arXiv:1408.5093 [cs.CV], 2014.

\bibitem{JiangMS2014}
Y.~Jiang, J.~Liu, and H.~Lu.
\newblock {Chat with illustration}.
\newblock {\em Multimedia Systems}, 2014.

\bibitem{JohnsonICCV2015}
J.~Johnson, A.~Karpathy, and L.~Fei-Fei.
\newblock Densecap: Fully convolutional localization networks for dense
  captioning.
\newblock {\em ICCV}, 2015.

\bibitem{JoshiTOMCCAP2006}
D.~Joshi, J.~Wang, and J.~Li.
\newblock {The Story Picturing Engine: a system for automatic text
  illustration}.
\newblock {\em ACM Transactions on Multimedia Computing, Communications, and
  Applications}, 2(1):68--89, 2006.

\bibitem{kalchbrenner14acl}
N.~Kalchbrenner, E.~Grefenstette, and P.~Blunsom.
\newblock A convolutional neural network for modelling sentences.
\newblock In {\em ACL}, 2014.

\bibitem{KalogerakisICCV2009}
E.~Kalogerakis, , O.~Vesselova, J.~Hays, A.~Efros, and A.~Hertzmann.
\newblock Image sequence geolocation with human travel priors.
\newblock In {\em ICCV}, pages 253--260, 2009.

\bibitem{karpathy15cvpr}
A.~Karpathy and L.~Fei-Fei.
\newblock Deep visual-semantic alignments for generating image descriptions.
\newblock In {\em CVPR}, 2015.

\bibitem{kim14emnlp}
Y.~Kim.
\newblock Convolutional neural networks for sentence classification.
\newblock In {\em EMNLP}, 2014.

\bibitem{KirosTACL2015}
R.~Kiros, R.~Salakhutdinov, and R.~Zemel.
\newblock Unifying visual-semantic embeddings with multimodal neural language
  models.
\newblock {\em TACL}, 2015.

\bibitem{KoppelLRE2011}
M.~Koppel, J.~Schler, and S.~Argamon.
\newblock Authorship attribution in the wild.
\newblock {\em Language Resources and Evaluation}, 45(1):83--94, 2011.

\bibitem{KovashkaIJCV2015}
A.~Kovashka, D.~Parikh, and K.~Grauman.
\newblock Whittlesearch: Interactive image search with relative attribute
  feedback.
\newblock {\em International Journal of Computer Vision}, pages 1--26, 2015.

\bibitem{krizhevsky12nips}
A.~Krizhevsky, I.~Sutskever, and G.~Hinton.
\newblock {ImageNet} classification with deep convolutional neural networks.
\newblock In {\em NIPS}, 2012.

\bibitem{KulkarniCVPR2011}
G.~Kulkarni, V.~Premraj, S.~Dhar, S.~Li, Y.~Choi, A.~Berg, and T.~Berg.
\newblock Baby talk: Understanding and generating image descriptions.
\newblock In {\em CVPR}. Citeseer, 2011.

\bibitem{KumarECCV2008}
N.~Kumar, P.~Belhumeur, and S.~Nayar.
\newblock Facetracer: A search engine for large collections of images with
  faces.
\newblock In {\em ECCV}, pages 340--353. Springer, 2008.

\bibitem{le14icml}
Q.~Le and T.~Mikolov.
\newblock Distributed representations of sentences and documents.
\newblock In {\em ICML}, 2014.

\bibitem{LiICM2011}
Z.~Li, M.~Wang, J.~Liu, C.~Xu, and H.~Lu.
\newblock {News contextualization with geographic and visual information}.
\newblock {\em International Conference on Multimedia}, page 133, 2011.

\bibitem{LinECCV2014}
T.~Lin, M.~Maire, S.~Belongie, J.~Hays, P.~Perona, D.~Ramanan, P.~Doll{\'a}r,
  and C.~Zitnick.
\newblock Microsoft coco: Common objects in context.
\newblock In {\em ECCV}, pages 740--755. 2014.

\bibitem{LongARXIV2015}
M.~Long and J.~Wang.
\newblock Learning transferable features with deep adaptation networks.
\newblock {\em arXiv preprint arXiv:1502.02791}, 2015.

\bibitem{maaten08tsne}
L.~Maaten and G.~Hinton.
\newblock Visualizing high dimensional data using t-sne.
\newblock {\em JMLR}, 9(11):2579--2605, 2008.

\bibitem{MaoICLR2015}
J.~Mao, W.~Xu, Y.~Yang, J.~Wang, Z.~Huang, and A.~Yuille.
\newblock Deep captioning with multimodal recurrent neural networks (m-rnn).
\newblock {\em ICLR}, 2015.

\bibitem{mardia79}
K.~Mardia, J.~Kent, and J.~Bibby.
\newblock {\em Multivariate Analysis}.
\newblock Academic Press, 1979.

\bibitem{mikolov13iclr}
T.~Mikolov, K.~Chen, G.~Corrado, and J.~Dean.
\newblock Efficient estimation of word representations in vector space.
\newblock In {\em ICLR}, 2013.

\bibitem{mikolov13nips}
T.~Mikolov, I.~Sutskever, K.~Chen, G.~Corrado, and J.~Dean.
\newblock Distributed representations of words and phrases and their
  compositionality.
\newblock In {\em NIPS}, 2013.

\bibitem{NarayananSP2012}
A.~Narayanan, H.~Paskov, N.~Gong, J.~Bethencourt, E.~Stefanov, E.~Shi, and
  D.~Song.
\newblock On the feasibility of internet-scale author identification.
\newblock In {\em Symposium on Security and Privacy}, pages 300--314. IEEE,
  2012.

\bibitem{OlivaPBR2006}
A.~Oliva and A.~Torralba.
\newblock Building the gist of a scene: The role of global image features in
  recognition.
\newblock {\em Progress in brain research}, 155:23--36, 2006.

\bibitem{OrdonezIJCV2015}
V.~Ordonez, X.~Han, P.~Kuznetsova, G.~Kulkarni, M.~Mitchell, K.~Yamaguchi,
  K.~Stratos, et~al.
\newblock Large scale retrieval and generation of image descriptions.
\newblock {\em International Journal of Computer Vision}, pages 1--14, 2015.

\bibitem{OrdonezNIPS2011}
V.~Ordonez, G.~Kulkarni, and T.~Berg.
\newblock Im2text: Describing images using 1 million captioned photographs.
\newblock In {\em NIPS}, pages 1143--1151, 2011.

\bibitem{PerlinCCGIT1996}
K.~Perlin and A.~Goldberg.
\newblock Improv: A system for scripting interactive actors in virtual worlds.
\newblock In {\em Conference on Computer Graphics and Interactive Techniques},
  pages 205--216. ACM, 1996.

\bibitem{PintoICWSD2013}
H.~Pinto, J.~Almeida, and M.~Gon{\c{c}}alves.
\newblock Using early view patterns to predict the popularity of youtube
  videos.
\newblock In {\em ACM International Conference on Web Search and Data Mining},
  pages 365--374, 2013.

\bibitem{QuattoniNAACL2016}
A.~Quattoni, A.~Ramisa, P.~Swaroop, E.~Simo-Serra, and F.~Moreno-Noguer.
\newblock Structured prediction with output embeddings for semantic image
  annotation.
\newblock In {\em Conference of the North American Chapter of the Association
  for Computational Linguistics}, 2016.

\bibitem{RashtchianNAACL2010}
C.~Rashtchian, P.~Young, M.~Hodosh, and J.~Hockenmaier.
\newblock Collecting image annotations using amazon's mechanical turk.
\newblock In {\em NAACL: Workshop on Creating Speech and Language Data with
  Amazon's Mechanical Turk}, pages 139--147, 2010.

\bibitem{RasiwasiaToM2007}
N.~Rasiwasia, P.~Moreno, and N.~Vasconcelos.
\newblock Bridging the gap: Query by semantic example.
\newblock {\em IEEE Transactions on Multimedia}, 9(5):923--938, 2007.

\bibitem{SerdyukovCRDIR2009}
P.~Serdyukov, V.~Murdock, and R.~Van Zwol.
\newblock Placing flickr photos on a map.
\newblock In {\em ACM Conference on Research and Development in Information
  Retrieval}, pages 484--491. ACM, 2009.

\bibitem{simonyan14arxiv}
K.~Simonyan and A.~Zisserman.
\newblock Very deep convolutional networks for large-scale image recognition.
\newblock arXiv:1409.1556 [cs.CV], 2014.

\bibitem{SocherCVPR2010}
R.~Socher and L~Fei-Fei.
\newblock Connecting modalities: Semi-supervised segmentation and annotation of
  images using unaligned text corpora.
\newblock In {\em CVPR}, pages 966--973. IEEE, 2010.

\bibitem{srivastava14jmlr}
N.~Srivastava, J.~Hinton, A.~Krizhevsky, I.~Sutskever, and R.~Salakhutdinov.
\newblock Dropout: A simple way to prevent neural networks from overfitting.
\newblock {\em JMLR}, 15(6):1929--1958, 2014.

\bibitem{StamatatosJASIS2009}
E.~Stamatatos.
\newblock A survey of modern authorship attribution methods.
\newblock {\em Journal of the American Society for Information Science and
  Technology}, 60(3):538--556, 2009.

\bibitem{TatarSNAM2014}
A.~Tatar, P.~Antoniadis, M.~De Amorim, and S.~Fdida.
\newblock From popularity prediction to ranking online news.
\newblock {\em Social Network Analysis and Mining}, 4(1):1--12, 2014.

\bibitem{TrampuSIKDD2012}
M.~Trampu{\v{s}} and B.~Novak.
\newblock Internals of an aggregated web news feed.
\newblock In {\em International Information Science Conference IS}, pages
  431--434, 2012.

\bibitem{TsagkiasAIR2010}
M.~Tsagkias, W.~Weerkamp, and M.~De Rijke.
\newblock News comments: Exploring, modeling, and online prediction.
\newblock In {\em Advances in Information Retrieval}, pages 191--203. Springer,
  2010.

\bibitem{VinyalsCVPR2015}
O.~Vinyals, A.~Toshev, S.~Bengio, and D.~Erhan.
\newblock Show and tell: A neural image caption generator.
\newblock In {\em CVPR}, pages 3156--3164, 2015.

\bibitem{WangACM2014}
Z.~Wang, P.~Cui, L.~Xie, W.~Zhu, Y.~Rui, and S.~Yang.
\newblock {Bilateral Correspondence Model for Words-and-Pictures Association in
  Multimedia-Rich Microblogs}.
\newblock {\em ACM Trans. Multimedia Computing, Communications, and
  Applications}, 10(4):1--21, 2014.

\bibitem{WeyandARXIV2016}
T.~Weyand, I.~Kostrikov, and James Philbin.
\newblock Planet - photo geolocation with convolutional neural networks.
\newblock arXiv:1602.05314 [cs.CV], 2016.

\bibitem{WingACL2011}
B.~Wing and J.~Baldridge.
\newblock Simple supervised document geolocation with geodesic grids.
\newblock In {\em Annual Meeting of the ACL: Human Language Technologies},
  pages 955--964. Association for Computational Linguistics, 2011.

\bibitem{XiaoCVPR2010}
J.~Xiao, J.~Hays, K.~Ehinger, A.~Oliva, and A.~Torralba.
\newblock Sun database: Large-scale scene recognition from abbey to zoo.
\newblock In {\em CVPR}, pages 3485--3492, 2010.

\bibitem{XuICML2015}
K.~Xu, J.~Ba, R.~Kiros, K.~Cho, A.~Courville, R.~Salakhudinov, R.~Zemel, and
  Y.~Bengio.
\newblock Show, attend and tell: Neural image caption generation with visual
  attention.
\newblock In {\em ICML}, 2015.

\bibitem{yan15cvpr}
F.~Yan and K.~Mikolajczyk.
\newblock Deep correlation for matching images and text.
\newblock In {\em CVPR}, 2015.

\bibitem{YosinskiNIPS2014}
J.~Yosinski, J.~Clune, Y.~Bengio, and H.~Lipson.
\newblock How transferable are features in deep neural networks?
\newblock In {\em NIPS}, pages 3320--3328, 2014.

\bibitem{YoungTACL2014}
P.~Young, A.~Lai, M.~Hodosh, and J.~Hockenmaier.
\newblock From image descriptions to visual denotations: New similarity metrics
  for semantic inference over event descriptions.
\newblock {\em Transactions of the Association for Computational Linguistics},
  2:67--78, 2014.

\bibitem{zhang14eccv}
Z.~Zhang, P.~Luo, C.~Loy, and X.~Tang.
\newblock Facial landmark detection by deep multitask learning.
\newblock In {\em ECCV}, 2014.

\bibitem{zhou14nips}
B.~Zhou, A.~Lapedriza, J.~Xiao, A.~Torralba, and A.~Oliva.
\newblock Learning deep features for scene recognition using places database.
\newblock In {\em NIPS}, 2014.

\bibitem{ZhouICM2012}
Y.~Zhou and J.~Luo.
\newblock {Geo-location inference on news articles via multimodal pLSA}.
\newblock {\em ACM International Conference on Multimedia}, page 741, 2012.

\bibitem{ZitnickICCV2013}
C.~Zitnick, D.~Parikh, and L.~Vanderwende.
\newblock Learning the visual interpretation of sentences.
\newblock In {\em ICCV}, pages 1681--1688, 2013.

\end{thebibliography}

\begin{IEEEbiography}[{\includegraphics[width=1in,height=1.25in,clip,keepaspectratio]{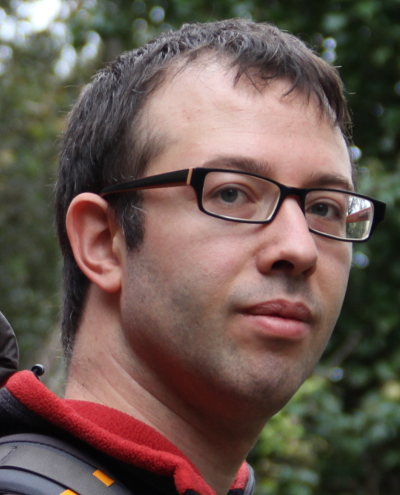}}]{Arnau Ramisa} received the MSc degree in computer science (computer vision) from the Autonomous University of Barcelona (UAB) in 2006, and in 2009 completed a PhD at the Artificial Intelligence Research Institute (IIIA-CSIC) and the UAB. From 2009 to the start of 2011, he was a postdoctoral fellow at the LEAR team in INRIA Grenoble / Rhone-Alpes, and since 2011 he is a research fellow at the Institut de Rob\`otica i Inform\`atica Industrial in Barcelona (IRI). His research interests include robot vision, object classification and detection, image retrieval and natural language processing.
\end{IEEEbiography}

\vspace{-1.1cm}

\begin{IEEEbiographynophoto}{Fei Yan} is a senior research fellow at University of Surrey in the United Kingdom. His research interests focus on machine learning, in particular kernel methods, structured learning, and deep neural networks. He is also interested in the application of machine learning to computer vision and natural language processing, such as object recognition, object tracking, natural language analysis and generation, and joint modelling of vision and language. He has publications in major machine learning and computer vision conferences and journals.
\end{IEEEbiographynophoto}

\vspace{-1.1cm}

\begin{IEEEbiography}[{\includegraphics[width=1in,height=1.25in,clip,keepaspectratio]{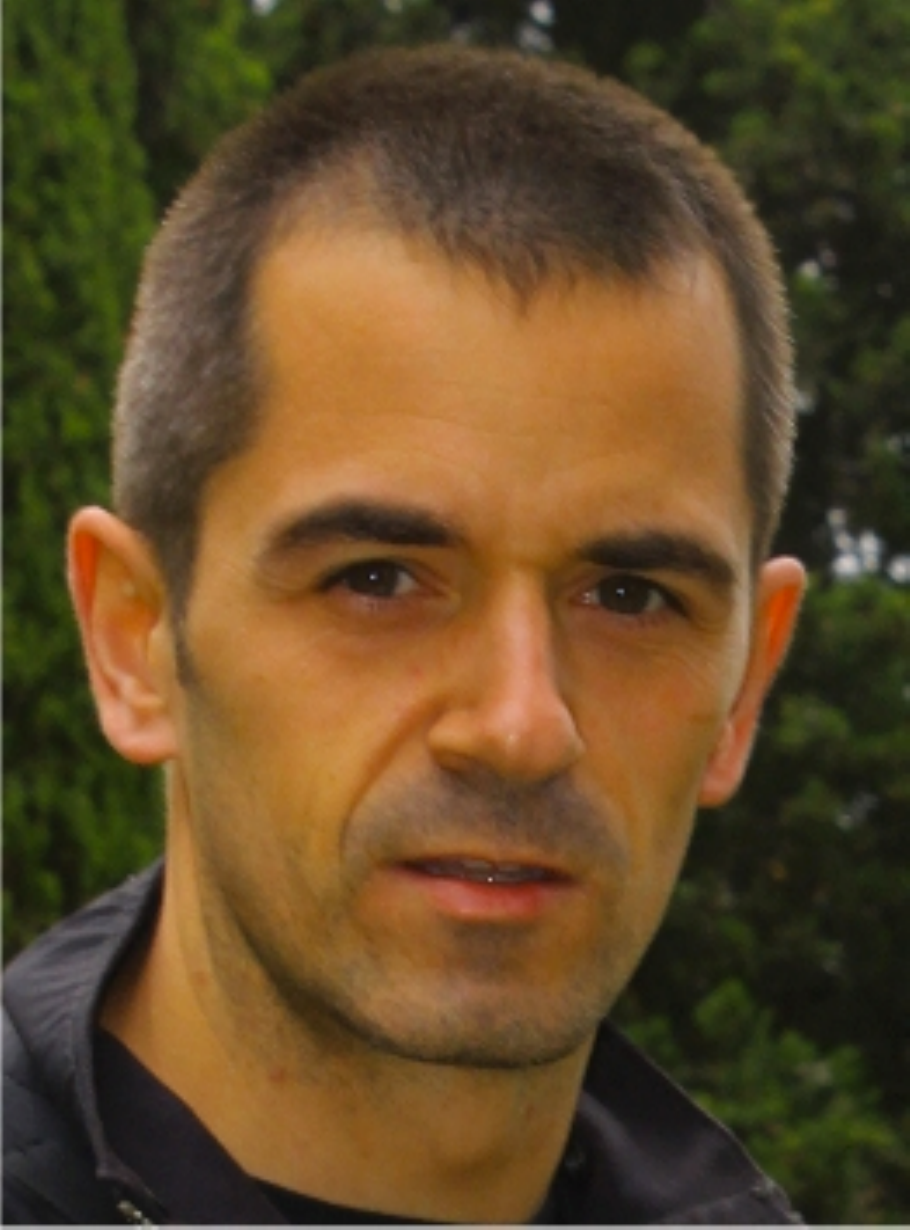}}]{Francesc Moreno-Noguer} received the MSc degrees in industrial engineering and electronics from the Technical University of Catalonia (UPC) and the Universitat de Barcelona in 2001 and
2002, respectively, and the PhD degree from UPC in 2005. From 2006 to 2008, he was a postdoctoral fellow at the computer vision departments of Columbia University and the \'Ecole Polytechnique F\'ed\'erale de Lausanne. In
2009, he joined the Institut de Rob\`otica i Inform\`atica Industrial in Barcelona as an associate researcher of the
Spanish Scientific Research Council. His research interests include
retrieving rigid and nonrigid shape, motion, and camera pose from single
images and video sequences. He received UPC's Doctoral Dissertation
Extraordinary Award for his work.
\end{IEEEbiography}

\vspace{-1.1cm}

\begin{IEEEbiography}[{\includegraphics[width=1in,height=1.25in,clip,keepaspectratio]{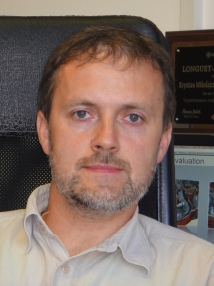}}]{Krystian Mikolajczyk}   is an Associate Professor at Imperial College London. He completed his PhD degree at the Institute National Polytechnique de Grenoble and held a number of research positions at INRIA, University of Oxford and Technical University of Darmstadt, as well as faculty positions at the University of Surrey, and Imperial College London. His main area of expertise is in image and video recognition, in particular methods for image representation and learning. He has served in various roles at major international conferences co-chairing British Machine Vision Conference 2012 and IEEE International Conference on Advanced Video and Signal-Based Surveillance 2013. In 2014 he received Longuet-Higgins Prize awarded by the Technical Committee on Pattern Analysis and Machine Intelligence of the IEEE Computer Society.  
\end{IEEEbiography}

%\begin{IEEEbiographynophoto}{Pascal Fua}

\end{document}